\pdfoutput=1

\documentclass[11pt]{article}

\usepackage[final]{acl}

\usepackage{times}
\usepackage{latexsym}

\usepackage[T1]{fontenc}

\usepackage[utf8]{inputenc}

\usepackage{microtype}

\usepackage{inconsolata}

\usepackage[hang,flushmargin]{footmisc}

\title{Confidence Estimation for LLMs in Multi-turn Interactions}
\author{
Caiqi Zhang\textsuperscript{1}\thanks{Equal contribution. Codes and data can be found in \href{https://github.com/caiqizh/multi-turn-conf.git}{GitHub}.}, 
Ruihan Yang\textsuperscript{2}\footnotemark[1], 
Xiaochen Zhu\textsuperscript{1},
Chengzu Li\textsuperscript{1}, 
Tiancheng Hu\textsuperscript{1}, \\
\textbf{Yijiang Dong}\textsuperscript{1},
\textbf{Deqing Yang}\textsuperscript{2},
\textbf{Nigel Collier}\textsuperscript{1} \\
\\
\textsuperscript{1}University of Cambridge \quad 
\textsuperscript{2}Fudan University \quad \\
\\
\texttt{\{cz391,nhc30\}@cam.ac.uk$^{1}$, \{rhyang17,yangdeqing\}@fudan.edu.cn$^{2}$,}
}
\usepackage{microtype}
\usepackage{graphicx}
\usepackage{arydshln}
\usepackage{inconsolata}
\usepackage{todonotes}
\usepackage{tabularx}
\usepackage{booktabs} 
\usepackage{multirow}
\usepackage{amsmath}
\usepackage{amssymb}
\usepackage{graphicx}
\usepackage{subcaption}
\usepackage{fge}

\usepackage[framemethod=TikZ]{mdframed} 
\usepackage{listings}
\usepackage{longtable}
\usepackage{tcolorbox}
\usepackage{xspace}
\usepackage{wrapfig}
\usepackage{pgfplots}
\usepackage{color, colortbl}
\usepackage{array}
\usepackage{paralist}
\usepackage{geometry}
\usepackage{enumitem}
\usepackage{float}  
\usepackage{makecell}
\usepackage{pifont} 
\usepackage{xcolor}  
\usepackage{colortbl}

\usepackage{hyperref}

\newcommand{\rparagraph}[1]{\vspace{1.2mm}\noindent\textbf{#1}}

\definecolor{Gray}{gray}{0.92}
\definecolor{racing-green}{rgb}{0.0, 0.8, 0.6}
\definecolor{awesome-red}{rgb}{1.0, 0.13, 0.32}
\definecolor{LightCyan}{rgb}{0.88,1,1}
\definecolor{darkgreen}{RGB}{0,150,0}
\definecolor{Ground}{RGB}{255,184,55}
\definecolor{Dirt}{RGB}{191,169,115}
\definecolor{Pink}{RGB}{226,184,176}
\definecolor{Violet}{RGB}{163,148,170}
\definecolor{darkred}{RGB}{150,0,0} 
\definecolor{Red}{RGB}{171, 61, 56}
\definecolor{Green}{RGB}{62, 139, 117}
\definecolor{Blue}{RGB}{48, 110, 184}

\definecolor{CC}{RGB}{198, 226, 212} 
\definecolor{UU}{RGB}{198, 228, 253} 
\definecolor{CU}{RGB}{247, 202, 193} 
\definecolor{UC}{RGB}{242, 224, 253}

\newcommand{\ie}{\textit{i}.\textit{e}.,\ }
\newcommand{\eg}{\textit{e}.\textit{g}.,\ }

\definecolor{level4}{RGB}{110,136,203}
\definecolor{level3}{RGB}{173,190,226}
\definecolor{level2}{RGB}{205,208,243}
\definecolor{level1}{RGB}{236,236,252}

\newcommand{\cellcolorval}[1]{
   \ifdim#1pt>100pt \cellcolor{level4!95}#1\relax
   \else
   \ifdim#1pt>90pt \cellcolor{level4!85}#1\relax
   \else\ifdim#1pt>80pt \cellcolor{level3!75}#1\relax
   \else\ifdim#1pt>70pt \cellcolor{level3!60}#1\relax
   \else\ifdim#1pt>60pt \cellcolor{level2!70}#1\relax
   \else\ifdim#1pt>50pt \cellcolor{level2!45}#1\relax
   \else\ifdim#1pt>40pt \cellcolor{level2!30}#1\relax
   \else\ifdim#1pt>30pt \cellcolor{level2!10}#1\relax
   \else\ifdim#1pt>20pt \cellcolor{level1!10}#1\relax
   \else \cellcolor{level1!0}#1\relax
   \fi\fi\fi\fi\fi\fi\fi\fi
}

\newcommand{\cdashlinelr}[1]{\cdashline{#1}}
\newcolumntype{g}{>{\columncolor{Ground!5.2}}c}
\newcolumntype{d}{>{\columncolor{cyan!6}}c}
\newcolumntype{f}{>{\columncolor{lime!6}}c}
\newcolumntype{v}{>{\columncolor{purple!6}}c}
\newcolumntype{u}{>{\cellcolorval}c}

\definecolor{lightblue}{RGB}{130,169,217}
\definecolor{green}{RGB}{29,177,0}

\definecolor{Gray}{gray}{0.92}
\definecolor{racing-green}{rgb}{0.0, 0.8, 0.6}
\definecolor{awesome-red}{rgb}{1.0, 0.13, 0.32}
\definecolor{LightCyan}{rgb}{0.88,1,1}
\definecolor{darkgreen}{RGB}{0,150,0}
\definecolor{Ground}{RGB}{255,184,55}
\definecolor{Dirt}{RGB}{191,169,115}
\definecolor{Pink}{RGB}{226,184,176}
\definecolor{Violet}{RGB}{163,148,170}
\definecolor{darkred}{RGB}{150,0,0} %
\definecolor{bluelight}{RGB}{240,240,255}
\definecolor{greenight}{RGB}{240,255,240}

\newcommand{\twentyq}{\textsc{20Q}\xspace}
\newcommand{\guess}{\textsc{Guess}\xspace}
\newcommand{\grace}{\textsc{Grace}\xspace}
\newcommand{\trickme}{\textsc{TrickMe}\xspace}

\begin{document}
\maketitle
\begin{abstract}
While confidence estimation is a promising direction for mitigating hallucinations in Large Language Models (LLMs), current research dominantly focuses on single-turn settings. The dynamics of model confidence in multi-turn conversations, where context accumulates and ambiguity is progressively resolved, remain largely unexplored. 
Reliable confidence estimation in multi-turn settings is critical for many downstream applications, such as autonomous agents and human-in-the-loop systems.
This work presents the first systematic study of confidence estimation in multi-turn interactions, establishing a formal evaluation framework grounded in two key desiderata: per-turn calibration and monotonicity of confidence as more information becomes available.
To facilitate this, we introduce novel metrics, including a length-normalized Expected Calibration Error (\emph{InfoECE}), and a new "Hinter-Guesser" paradigm for generating controlled evaluation datasets. Our experiments reveal that widely-used confidence techniques struggle with calibration and monotonicity in multi-turn dialogues. 
We propose \textsc{P(Sufficient)}, a logit-based probe that achieves comparatively better performance, although the task remains far from solved. Our work provides a foundational methodology for developing more reliable and trustworthy conversational agents.
\end{abstract}

\section{Introduction}

Large Language Models (LLMs) have shown remarkable capabilities in multi-turn dialogue, collaborating with users on complex tasks \citep{wang2023mint, yi2024survey, laban2025llms}. Yet their tendency to “hallucinate” (\ie producing incorrect statements with high apparent certainty) remains a major obstacle for high-stakes use \citep{manakul2023selfcheckgpt,  zhang-etal-2024-need, shelmanov-etal-2025-head, hu2025navigating}. Confidence estimation, which aims to predict the likelihood that a model’s answer is correct, has accordingly become a promising direction for identifying and mitigating such failures \citep{zhang2025reinforcement, yang-etal-2025-logu, yang-etal-2025-uncle}.

\begin{figure}[t!]
    \centering    \includegraphics[width=0.9\linewidth]{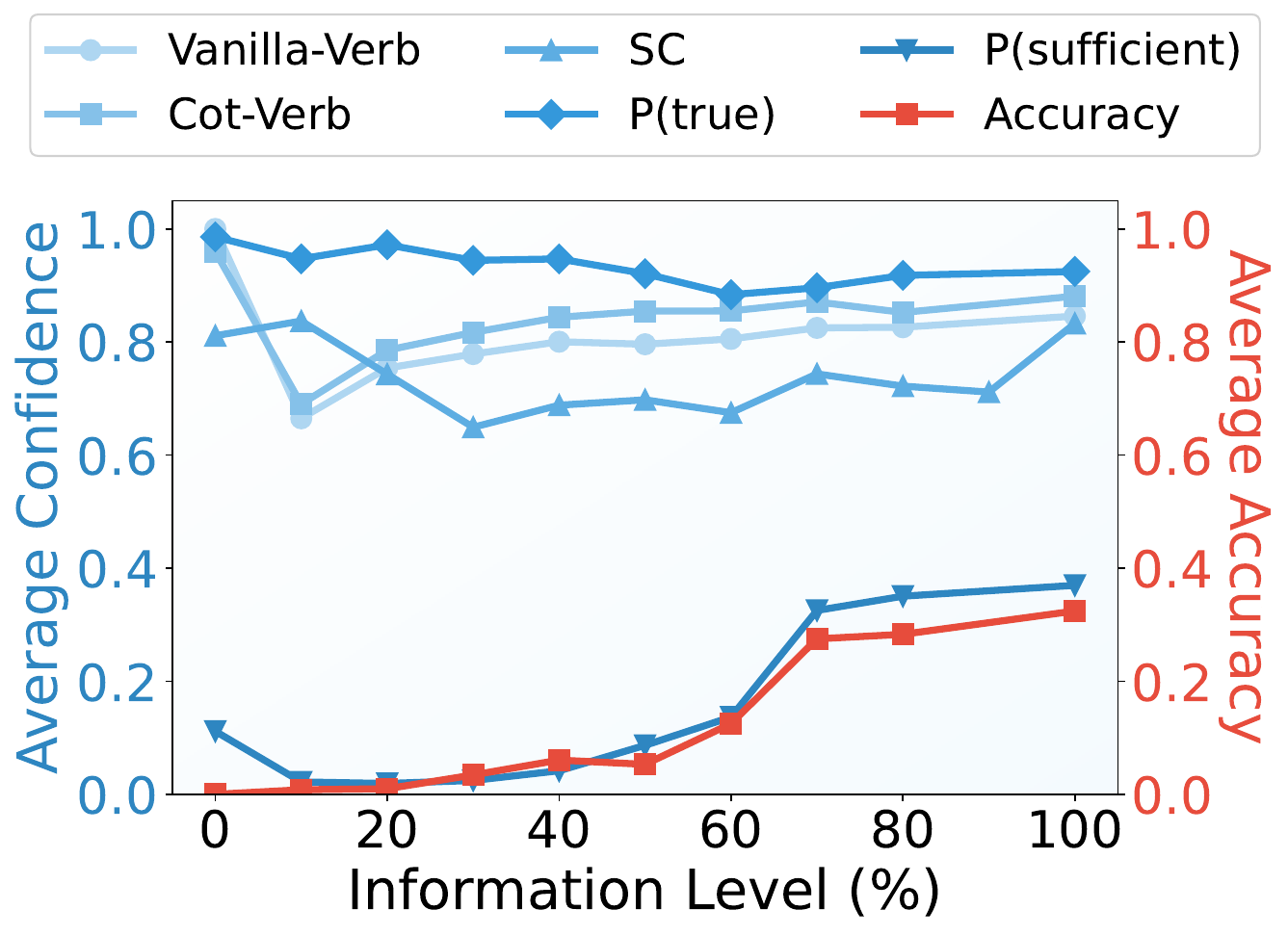}
    \caption{InfoECE on \guess (Llama3.1-70B). Ideally, confidence (\textcolor{cyan}{blue curves}) increases as more information is provided. Calibration improves when the confidence curves (\textcolor{cyan}{blue}) are closer to the accuracy curve (\textcolor{red}{red}). In this setting, \textsc{P(Sufficient)} best satisfies both monotonicity and calibration.} 
    \vspace{-2mm}
    \label{fig:head}
\end{figure}

Despite recent progress, most prior work studies confidence in single-turn question answering \citep{tian-etal-2023-just, xiong2023llms}, a static setup that overlooks the inherently \textbf{dynamic nature} of real human–AI interaction. 
In multi-turn conversations, information arrives incrementally: users refine their queries, models ask clarifying questions, and the hypothesis space narrows turn by turn. In such settings, confidence should not be a fixed attribute of a solitary response but a signal that evolves with the dialogue—ideally increasing as ambiguity is resolved and evidence accumulates. 
Reliable confidence estimation in this progression is therefore critical, as it serves as a decision-making heuristic for when to ask clarifying questions, invoke tools, or commit to actions in agentic workflows and human–AI collaboration. However, \textit{how well current methods track this progression is largely unknown}.

To address this gap, we present the first systematic study of confidence estimation in multi-turn conversations. We introduce a novel evaluation framework in which the model receives progressively more task-relevant information. We argue that in this controlled setting, a reliable confidence signal should satisfy two desiderata:  (1) \textbf{Calibration}, where the confidence accurately reflects empirical correctness at any given turn, and (2) \textbf{Monotonicity}, where confidence consistently increases as more useful information becomes available.

Guided by these desiderata, we develop new metrics and datasets tailored to confidence estimation in multi-turn settings. To measure calibration across dialogues of varying lengths, we introduce a length-normalized Expected Calibration Error at information level (\emph{InfoECE}). To quantify monotonicity, we employ Kendall’s $\tau$, a non-parametric rank correlation coefficient. We establish evaluation testbeds for two distinct regimes: (1) \textit{under-specified initial queries}, for which we introduce a novel ``Hinter--Guesser" paradigm to generate dialogues with progressively revealed clues; and (2) \textit{difficult but fully-specified queries}, for which we adapt existing incremental QA benchmarks \citep{wallace-etal-2019-trick, sung-etal-2025-grace} that provide sequential hints toward the correct answer.

Our experiments evaluate a suite of confidence estimation methods across four open-source models (\S~\ref{sec:setup}), revealing key insights into their multi-turn behavior. 
\textbf{First} (\S~\ref{sec:reliable-multiturn}), we find that widely used techniques struggle to maintain calibration or exhibit consistent monotonicity as conversations progress, as illustrated in Figure \ref{fig:head}. 
Our proposed method, \textsc{P(Sufficient)}, proves comparatively more effective; however, substantial room for improvement remains.
Meanwhile, we find that models exhibit stronger monotonicity when confidence is evaluated against the ground-truth answer rather than the model’s provisional answer at each turn. 
\textbf{Second} (\S~\ref{sec:info-vs-length}), we examine whether confidence increases are driven by added information or merely by turn count. \textsc{P(Sufficient)} more effectively distinguishes meaningful information gains from conversational filler.
\textbf{Finally} (\S~\ref{sec:single-vs-multi}), our analysis reveals that while model accuracy is comparable between multi-turn dialogues and single-turn summaries, the confidence signals behave very differently, underscoring that the interactive structure of the dialogue is crucial to models' confidence estimation. 
Overall, our findings highlight multi-turn confidence as a distinct and necessary target for reliable, decision-oriented LLM behavior.

\section{Related Work}

\rparagraph{Confidence Estimation in LLMs} 
Confidence and uncertainty estimation has been extensively studied in LLMs \citep{geng-etal-2024-survey}. More specifically, uncertainty reflects the variability in the model’s predictions given \textit{only the input query}, while confidence is defined with respect to \textit{both the input and the specific generated output}, indicating how certain the model is about that particular response~\citep{lin2023generating, zhang-etal-2024-luq, zhang2025grace}. 
Mainstream confidence estimation approaches include prompting-based (verbalized) methods~\citep{tian-etal-2023-just, dong-etal-2024-llm}, consistency-based methods~\citep{manakul2023selfcheckgpt, zhang-etal-2024-luq}, and logit-based methods~\citep{kadavath2022language}. These methods have been applied to various tasks, such as short-form factual QA~\citep{tian-etal-2023-just, lin2023generating}, long-form factual QA~\citep{zhang-etal-2024-luq, zhang2024atomic, zhang2025reinforcement}, and reasoning tasks~\citep{zhang2025roads, zhang-zhang-2025-cot}. 
However, a major limitation of existing works is their focus on single-turn settings. The effectiveness of confidence estimation in multi-turn conversations remains underexplored~\citep{kirchhof2025position}, where model confidence may \textbf{evolve dynamically} throughout the interaction. 
Our work aims to fill this gap by systematically evaluating existing confidence estimation methods and proposing novel approaches in multi-turn contexts.

\rparagraph{LLMs in Multi-turn Interactions} 
There has been growing interest in studying LLMs in multi-turn scenarios~\citep{laban2025llms, zhu-etal-2025-conformity}. Modern LLMs support interactive dialogue, enabling users to collaborate with the model across multiple turns to accomplish complex tasks. However, recent studies show that LLMs often perform significantly worse on the same tasks when framed in a multi-turn context compared to a single-turn setting~\citep{laban2025llms}. 
\citet{laban2025llms} also point out that many prior works~\citep{bai-etal-2024-mt, kwan-etal-2024-mt, duan-etal-2024-botchat} simulate \textit{episodic conversations}, where each turn introduces a subtask related to previous turns but can be evaluated in isolation. Under this framing, multi-turn tasks differ structurally from single-turn tasks and are not evaluated on the same set of questions. \citet{laban2025llms} argue that episodic tasks tend to overestimate LLM performance in multi-turn settings and construct a \textit{sharded} data construction method. 
In our work, we follow a similar sharded question construction strategy. For each question, we create multiple variants with increasing levels of contextual information provided across turns. This allows us to directly compare confidence estimation methods under varying levels of complexity.

Concurrent work \citep{zhang2026confidence} addresses multi-turn calibration under adversarial persuasion, where confidence should resist misleading user feedback. We study the complementary cooperative regime, where confidence should rise monotonically as information accumulates.

\section{Methodology}

\subsection{Notations}
We study confidence estimation in \emph{multi-turn} dialogue between a \textit{user} and an \textit{LLM}.
Dialogs are indexed by $d\in\{1,\dots,N\}$ and have $L_d$ turns, where $L_d$ \emph{may differ across dialogs}.
At turn $i\in\{1,\dots,L_d\}$, let the dialogue history be
\[
h_{d,i}=\{q_{d,1},a_{d,1},\dots,q_{d,i-1},a_{d,i-1}\}.
\]
The turn-$i$ prompt consists of the task description $T$ and the history $h_{d,i}$.
Model $M$ returns an answer $\hat y_{d,i}$ and a confidence $c_{d,i}\in[0,1]$.
Each dialog has one gold label $y_d$, and we record correctness
\[
z_{d,i}=\mathbb{I}[\hat y_{d,i}=y_d].
\]
\rparagraph{Task characteristics.} 
We design a controlled task that should exhibit three key properties:
\textit{C1: Progressive Information Acquisition.} Each turn reveals additional task-relevant information that narrows the hypothesis space or supports step-by-step reasoning.
\textit{C2: Step-wise Answerability and Evaluation.} At every turn the model outputs an answer and a confidence, enabling per-turn accuracy and calibration assessment.
\textit{C3: Monotonic Confidence Progression.} Confidence should increase with the turn index and align more closely with true accuracy, providing a usable reliability signal.
These properties yield a controllable testbed in which information strictly accumulates across turns, avoiding the limitations of episodic, non-progressive interactions.


\subsection{Two Initial-Question Regimes}

Based on the task characteristics, we consider two regimes defined by the completeness of the initial question.
\textbf{(1) Under-specified:} The initial question $q_{d,1}$ admits many plausible answers. Hints progressively \emph{prune} the candidate set.
\textbf{(2) Fully-specified but difficult:} The initial question $q_{d,1}$ pinpoints a unique answer in principle, but is hard to answer due to the models' knowledge limitation or reasoning ability. Hints make solving \emph{easier}.


\subsection{Evaluation Protocol and Metrics}

As a dialogue progresses, later turns include at least as much information as earlier ones.
Therefore, a useful confidence signal should have at least:
\begin{enumerate}[label=\alph*), topsep=1pt, itemsep=1pt]
  \item \textit{Per-level calibration:} within the same (normalized) information level, average confidence matches empirical accuracy.
  \item \textit{Monotonicity:} typically $c_{d,i+1}\ge c_{d,i}$;
\end{enumerate}

Because dialogue lengths vary, we first normalize turn \(i\) of dialogue \(d\) to a fractional information level
\[
s_{d,i}=\frac{i}{L_d}\in(0,1].
\]
where \(L_d\) is the number of turns in dialogue \(d\).
We then partition \([0,1]\) into \(B\) bins \(\{S_b\}_{b=1}^{B}\) (either equal-width or equal-mass) and index all turn positions by
$
\mathcal{I} = \{(d,i): 1\le i\le L_d\}.
$
The subset of indices whose normalized level falls into bin \(b\) is
$
\mathcal{I}_b = \{(d,i)\in\mathcal{I}: s_{d,i}\in S_b\}.
$
For each information level \(b\), the average confidence and accuracy are

\begingroup\footnotesize
\[
\mathrm{conf}_b = \frac{1}{|\mathcal{I}_b|}\sum_{(d,i)\in\mathcal{I}_b} c_{d,i},
\qquad
\mathrm{acc}_b  = \frac{1}{|\mathcal{I}_b|}\sum_{(d,i)\in\mathcal{I}_b} z_{d,i},
\]
\endgroup

where \(c_{d,i}\in[0,1]\) is the model’s per-turn confidence and \(z_{d,i}\in\{0,1\}\) indicates correctness of the turn-\(i\) answer in dialogue \(d\).

\rparagraph{Information-level ECE (InfoECE)} 
We compute an information-level ECE that groups predictions by normalized information exposure, enabling fair calibration comparisons across dialogues of different lengths.

\begingroup\footnotesize
\[
\mathrm{InfoECE}
=\frac{1}{B}\sum_{b=1}^{B}\big|\mathrm{acc}_b-\mathrm{conf}_b\big|.
\]
\endgroup


\rparagraph{Kendall’s $\tau$.}
Kendall’s $\tau$ measures the pairwise monotonic trend of confidence over turns.
For dialog $d$, consider all $\binom{L_d}{2}$ pairs $(i<j)$: a pair is \emph{concordant} if $c_{d,j}>c_{d,i}$ and \emph{discordant} if $c_{d,j}<c_{d,i}$ (ties ignored).

\begingroup\footnotesize
\[
\tau^{(d)}=\frac{N^{(d)}_{\mathrm{con}}-N^{(d)}_{\mathrm{dis}}}{\binom{L_d}{2}},\quad
\overline{\tau}=\frac{1}{N}\sum_{d=1}^{N}\tau^{(d)}.
\]
\endgroup
Values lie in $[-1,1]$: $1$ means strictly increasing confidences, and $0$ no overall trend.

\subsection{Confidence Estimation Methods}

Given the diversity of confidence estimation methods, we focus on the following three representative categories. We explicitly exclude post-hoc calibration techniques in this study (\eg Platt scaling and temperature scaling), as they represent orthogonal research directions, which aim to rescale the confidence scores using statistical procedures \citep{zhou2025beyond, zhang2024atomic}.

\rparagraph{Verbalized confidence}
We apply two verbalized prompting strategies \citep{tian-etal-2023-just} to elicit confidence scores directly from the model (see prompts in Appendix~\ref{app:ambig_prompt}):

\begin{enumerate}[label=\it\alph*), topsep=2pt, itemsep=1pt]
\item \textsc{Vanilla-Verb:} Given a candidate answer, the model is required to self-report confidence in $[0,100]$; rescale to $[0,1]$ for $c_{d,i}$.
\item \textsc{CoT-Verb:} Different from \textsc{Vanilla-Verb}, the model is now required to think step by step before given the self-reported confidence in $[0,100]$; rescale to $[0,1]$.
\end{enumerate}

\rparagraph{Self-consistency (SC)}
Given the question, we independently sample $m$ (\eg 20) answers $a^{(1)}_{d,i},\dots,a^{(m)}_{d,i}$.
For any answer $a$, we define the confidence as the fraction of samples that match $a$:
\begin{equation*}
\begin{aligned}
& c_{d,i}=\frac{1}{m}\sum_{j=1}^{m}\mathbb{I}[a^{(j)}_{d,i}=a].
\end{aligned}
\end{equation*}

\rparagraph{Logit-based probes}
We leverage internal model signals to estimate prediction confidence (see prompts in Appendix~\ref{app:ambig_prompt}). (1) \textsc{P(True)} \citep{kadavath2022language}: At step $i$, given the prompt $p_i$, we first elicit the answer $a_i$ from model $M$. We force a binary choice \textbf{A.~True} vs.\ \textbf{B.~False} with output constrained to a single uppercase letter. The confidence score is the model’s softmax probability assigned to label \textbf{A}.

Unlike \textsc{P(True)}, which asks if the answer is correct, we propose a new method that probes the confidence by asking model if the current information is \textbf{sufficient} (\textsc{P(Sufficient)}) to entail that answer $a$ is the only correct answer. \textsc{P(Sufficient)} works particular well in our under-specified settings, where the set of plausible answers shrinks with each turn. This method allows the model to express low confidence even if its current guess happens to be correct, as long as other candidates have not yet been ruled out by the provided hints. This aligns the confidence score more closely with the true identifiability of the answer from the accumulated evidence, rather than mere incidental correctness.

Set $c_{d,i}=P_T(d,i)$ or $c_{d,i}=P_S(d,i)$, we ask the model two binary probes about $\hat y_{d,i}$ under $p_{d,i}$:
\[
\begin{aligned}
P_T(d,i) &= \Pr[\texttt{A}\mid p_{d,i}, \hat y_{d,i};~\textsc{P(True)}],\\
P_S(d,i) &= \Pr[\texttt{A}\mid p_{d,i}, \hat y_{d,i};~\textsc{P(Sufficient)}].
\end{aligned}
\]

\begin{table*}[t]
\setlength\tabcolsep{11pt}
\scriptsize
  \centering
    \begin{tabularx}{0.95\linewidth}{p{0.8cm}|X}
    \toprule
    \textbf{Dataset} & \textbf{Exemplar Prompt at Turn 4} \\
    \midrule
    \multirow{7}{*}{\textbf{\twentyq}} & \textbf{User:} Given the following information, provide the title of the Wikipedia page that best answers the last question fragment. If unsure, provide your best guess. The answer should be concise. You have some clues about the answer: \\
    & \textbf{Assistant:} Is it human-made? \textbf{User:} Yes \\
    & \textbf{Assistant:} Is it typically found indoors? \textbf{User:} Yes \\
    & \textbf{Assistant:} Is it commonly encountered in living rooms? \textbf{User:} Yes \\
    & \textbf{Assistant:} Is it larger than a book? \textbf{User:} Yes \\
    & \textbf{User:} Now guess what it is: \\
    & \textit{Keyword:} \colorbox{blue!15}{\textcolor{blue}{\textbf{television}}} \\
    \midrule
    \multirow{7}{*}{\parbox{2cm}{\textbf{\guess}}} & \textbf{User:} Given the following information, name the single CITY that best fits them. If unsure, provide your best guess. The answer should be concise. You have some clues about the answer: \\
    & \textbf{Assistant:} What continent is the city in? \textbf{User:} Asia \\
    & \textbf{Assistant:} Is the city coastal or inland? \textbf{User:} Inland \\
    & \textbf{Assistant:} What's the climate like in the city? \textbf{User:} Tropical \\
    & \textbf{Assistant:} What region within the continent is the city located? \textbf{User:} Southeast Asia \\
    & \textbf{User:} Now guess what it is: \\
    & \textit{Keyword:} \colorbox{blue!15}{\textcolor{blue}{\textbf{Bogor, Indonesia}}} \\
    \bottomrule
    \end{tabularx}
    \caption{Examples from the \twentyq and \guess datasets. Ideally, the model's confidence should increase monotonically throughout the conversation. Note that the conversation history is collected from our Hinter-Guesser pipeline and remains fixed for confidence evaluation (\ie no strategic information-gathering is involved during evaluation).}  \label{tab:main_examples}
\end{table*}

\section{Dataset Construction}

For \emph{under-specified} (initially many plausible answers) regime, we construct our own datasets with \twentyq and \guess (as shown in Table \ref{tab:main_examples}). For \emph{fully-specified} (a unique answer exists from the start but is hard to infer until sufficient evidence accumulates) regime, we directly apply existing datasets: \grace \citep{sung-etal-2025-grace} and \trickme \citep{wallace-etal-2019-trick} (examples in Table \ref{tab:complete_prompts} in Appendix).

\subsection{Under-specified Datasets}
We primarily leverage 20Q and Guess-my-City (\guess)-style \citep{abdulhai2023lmrl} settings for the under-specified regime. In both, an answerer holds a secret entity (an entity for \twentyq; a city for \guess). The questioner incrementally seeks information to recover the secret entity. The key difference is that \twentyq constrains the questioner to yes/no questions, whereas \guess permits open-ended questions. This incremental, information-seeking interaction naturally satisfies \textbf{C1}.
Crucially, the setting also enables \textbf{C2}: at every turn the questioner can issue a concrete guess, even when information is still incomplete. This contrasts with math problem settings (\eg \citet{laban2025llms}), where intermediate turns often lack the conditions required to score correctness, impeding per-turn accuracy assessment.
However, naively simulating two LLMs to play the roles of questioner and answerer can \textbf{violate} \textbf{C3}. Early turns may contain irrelevant or misleading questions, yielding stagnant or even decreasing confidence. 
To address this, we reformulate the interaction into a \emph{Hinter–Guesser} paradigm that structures the information flow while retaining uncertainty. 

\rparagraph{Hinter–Guesser Paradigm}
(1) \textbf{QA Stage.} A \emph{Hinter} (LLM) is assigned a secret entity and must provide, at each turn, a helpful but non-trivial hint. A \emph{Guesser} then (i) makes a best-guess answer and (ii) flags whether multiple answers remain plausible (\emph{uniqueness probing}).
(2) \textbf{Uniqueness Probing.} Even when the Guesser’s answer is correct, the Guesser indicates if other candidates still fit the evidence. This distinguishes a coincidentally right guess from the moment the answer becomes suficiently supported by the clues, aligning confidence with identifiability rather than chance.
(3) \textbf{Stopping \& Filtering.} The dialogue proceeds until the Guesser both answers correctly and certifies uniqueness. We retain only successful dialogues (eventually solvable) and discard trajectories that fail to converge. 
In total, we collected 1,848 dialogue turns from \twentyq, spanning 226 entities. For \guess, we collected 1,625 turns spanning 223 entities. Then during confidence evaluation, since the conversations are \textbf{fixed}, models \textbf{do not} perform any strategic decision-making; instead, they only make guesses and predict confidence scores.


\subsection{Fully-specified Datasets}
We select two incremental, quizbowl-style QA datasets: \grace \citep{sung-etal-2025-grace} and \trickme \citep{wallace-etal-2019-trick}, where clues become increasingly specific and a unique gold answer exists from the outset. \grace and \trickme directly supporting \textbf{C1–C3} as evidence strengthens. We follow their standard protocols without modification, report per-turn accuracy and confidence, and defer details to the Appendix \ref{app:incremental_qa}.

\newcolumntype{B}{>{\columncolor{lightblue!5}}c}
\newcolumntype{G}{>{\columncolor{green!6}}c}
\newcolumntype{P}{>{\columncolor{purple!2.4}}c}
\newcolumntype{A}{>{\columncolor{bluelight!32}}c}
\newcolumntype{C}{>{\columncolor{greenight!32}}c}

\begin{table*}[t]
\centering
\footnotesize
\setlength\tabcolsep{14pt}
\scalebox{0.75}{ 
\begin{tabular}{llAACCPPBB} 
\toprule
 & \multirow{2}{*}{\textbf{Method}} 
& \multicolumn{2}{c}{\textbf{\twentyq}} 
& \multicolumn{2}{c}{\textbf{\guess}} 
& \multicolumn{2}{c}{\textbf{\grace}} 
& \multicolumn{2}{c}{\textbf{\trickme}} \\
\cmidrule(lr){3-4} \cmidrule(lr){5-6} \cmidrule(lr){7-8} \cmidrule(lr){9-10}
& & \cellcolor{gray!0}InfoECE & \cellcolor{gray!0}$\tau$ & \cellcolor{gray!0}InfoECE & \cellcolor{gray!0}$\tau$ & \cellcolor{gray!0}InfoECE & \cellcolor{gray!0}$\tau$ & \cellcolor{gray!0}InfoECE & \cellcolor{gray!0}$\tau$ \\
\midrule
\multirow{6}{*}{\rotatebox{90}{\textbf{LLama3.1-8b}}}
& {\cellcolor{gray!5}\textit{\textbf{Accuracy}}} & \multicolumn{2}{c}{\cellcolor{gray!5}\textit{\textbf{24.95}}} & \multicolumn{2}{c}{\cellcolor{gray!5}\textit{\textbf{14.52}}} & \multicolumn{2}{c}{\cellcolor{gray!5}\textit{\textbf{35.73}}} & \multicolumn{2}{c}{\cellcolor{gray!5}\textit{\textbf{41.91}}} \\
\cdashlinelr{3-10}
& \textsc{Vanilla-Verb} & 67.82 & -6.36 & 74.89 & -6.58 & 51.00 & 52.21 & 59.26 & 54.55 \\
& \textsc{Cot-Verb} & 63.75 & \textbf{46.97} & 70.28 & 37.84 & 45.21 & 43.25 & 87.70 & 48.63 \\
& \textsc{SC} & \textbf{18.05} & 36.73 & 38.14 & 9.43 & \textbf{10.57} & 52.40 & \textbf{18.97} & 55.37 \\
& \textsc{P(true)} & 69.02 & 42.10 & 67.08 & 19.91 & 50.43 & 48.48 & 55.61 & 52.37 \\
& \textsc{P(sufficient)} & 41.08 & 38.57 & \textbf{35.17} & \textbf{68.51} & 23.77 & \textbf{53.94} & 33.74 & \textbf{58.34} \\
\midrule
\multirow{6}{*}{\rotatebox{90}{\textbf{Llama3.1-70b}}}
& {\cellcolor{gray!5}\textit{\textbf{Accuracy}}} & \multicolumn{2}{c}{\cellcolor{gray!5}\textit{\textbf{33.87}}} & \multicolumn{2}{c}{\cellcolor{gray!5}\textit{\textbf{18.58}}} & \multicolumn{2}{c}{\cellcolor{gray!5}\textit{\textbf{48.27}}} & \multicolumn{2}{c}{\cellcolor{gray!5}\textit{\textbf{53.75}}} \\
\cdashlinelr{3-10}
& \textsc{Vanilla-Verb} & 59.63 & 17.60 & 65.52 & 16.92 & 39.06 & 47.13 & 47.47 & 44.49 \\
& \textsc{Cot-Verb} & 58.39 & 34.49 & 70.16 & 18.24 & 96.04 & 61.30 & 80.97 & 57.27 \\
& \textsc{SC} & 32.99 & 28.98 & 56.88 & 2.59 & 15.91 & 41.36 & \textbf{19.90} & 38.26 \\
& \textsc{P(true)} & 67.82 & 40.82 & 79.97 & 3.29 & 37.04 & 58.94 & 35.62 & 64.25 \\
& \textsc{P(sufficient)} & \textbf{13.05} & \textbf{48.43} & \textbf{5.27} & \textbf{81.51} & \textbf{11.52} & \textbf{66.86} & 23.16 & \textbf{71.38} \\
\midrule
\multirow{6}{*}{\rotatebox{90}{\textbf{Qwen2.5-7b}}}
& {\cellcolor{gray!5}\textit{\textbf{Accuracy}}} & \multicolumn{2}{c}{\cellcolor{gray!5}\textit{\textbf{25.22}}} & \multicolumn{2}{c}{\cellcolor{gray!5}\textit{\textbf{12.92}}} & \multicolumn{2}{c}{\cellcolor{gray!5}\textit{\textbf{27.34}}} & \multicolumn{2}{c}{\cellcolor{gray!5}\textit{\textbf{34.06}}} \\
\cdashlinelr{3-10}
& \textsc{Vanilla-Verb} & 58.05 & \textbf{61.13} & 48.68 & 26.99 & 50.40 & \textbf{55.55} & 54.62 & \textbf{56.39} \\
& \textsc{Cot-Verb} & 64.93 & 52.60 & 71.84 & \textbf{56.01} & 66.20 & 50.38 & 65.37 & 49.89 \\
& \textsc{SC} & 45.44 & 13.85 & 50.53 & 36.10 & 32.78 & 40.16 & \textbf{33.28} & 42.54 \\
& \textsc{P(true)} & 46.68 & 47.16 & 37.86 & 22.04 & 35.15 & 44.04 & 39.45 & 51.00 \\
& \textsc{P(sufficient)} & \textbf{36.64} & 55.24 & \textbf{26.63} & 51.44 & \textbf{28.67} & 47.79 & 35.26 & 52.11 \\
\midrule
\multirow{6}{*}{\rotatebox{90}{\textbf{Qwen2.5-72b}}}
& {\cellcolor{gray!5}\textit{\textbf{Accuracy}}} & \multicolumn{2}{c}{\cellcolor{gray!5}\textit{\textbf{32.36}}} & \multicolumn{2}{c}{\cellcolor{gray!5}\textit{\textbf{16.12}}} & \multicolumn{2}{c}{\cellcolor{gray!5}\textit{\textbf{47.49}}} & \multicolumn{2}{c}{\cellcolor{gray!5}\textit{\textbf{53.88}}} \\
\cdashlinelr{3-10}
& \textsc{Vanilla-Verb} & 47.92 & 67.97 & 67.04 & 52.81 & 43.18 & 72.59 & 41.62 & \textbf{71.32} \\
& \textsc{Cot-Verb} & 51.43 & \textbf{72.33} & 64.63 & 79.00 & 46.49 & \textbf{73.04} & 43.50 & 70.35 \\
& \textsc{SC} & 45.69 & 28.90 & 68.93 & 12.52 & 32.38 & 49.17 & 36.50 & 48.52 \\
& \textsc{P(true)} & \textbf{42.12} & 68.88 & 57.56 & 54.87 & \textbf{31.86} & 64.02 & 32.87 & 69.28 \\
& \textsc{P(sufficient)} & 45.86 & 66.81 & \textbf{28.32} & \textbf{83.76} & 32.93 & 66.04 & \textbf{32.41} & 71.24 \\
\bottomrule
\end{tabular}
}
\caption{InfoECE and $\tau$ across models and datasets. Numbers are in percentages and best results are \textbf{bolded}.}
\label{tab:main}
\end{table*}

\section{Experiments}
\label{sec:setup}
\rparagraph{Models.}
In our experiments, we use Llama3.1 Instruct (8B and 70B)~\citep{llama3modelcard}, Qwen2.5 Instruct (8B and 72B)~\citep{yang2024qwen2technicalreport}. Temperature is set to 1 for sampling and otherwise 0. 

\rparagraph{Confidence Estimation.}
For a fair comparison across methods and models, we first let the model answer the question once to obtain an answer $a$. We compare $a$ with the ground truth answer and label it correct or not. 
For each confidence estimation method, we then estimate the model confidence in this answer $a$; in parallel, we also estimate the model's confidence at each turn with respect to the ground-truth answer.

\rparagraph{Controlling for Conversational Length.}
A core hypothesis of our work is that a reliable confidence signal should increase monotonically as more task-relevant information becomes available. 
However, this trend could be a superficial artifact of dialogue length, where models become more confident simply because the turn index $i$ is higher. 
To disentangle these factors, we design the following experiment:
For a given dialogue $d$ at turn $i$, we create an adversarial condition by replacing the original informative hint with a \textbf{placebo QA pair} that adds conversational history without revealing task-relevant information (\eg Q: ``Is this a valid hint?'' A: ``Yes.''). We then compare the model accuracy and confidence across three states:
\begin{enumerate}[topsep=2pt, itemsep=1pt]
    \item \textbf{Baseline (turn $i-1$):} The model's prediction and confidence given history $h_{d,i-1}$.
    \item \textbf{Original (turn $i$):} Prediction and confidence after processing the original informative hint from turn $i$.
    \item \textbf{Placebo (turn $i'$):} Prediction and confidence after processing $h_{d,i-1}$ followed by the uninformative placebo hint.
\end{enumerate}
If confidence methods are robustly tracking information, we expect a significant increase in accuracy and confidence from the baseline to the original state. 
In contrast, the transition from the baseline to the adversarial state should yield a negligible change, despite the additional turn.

\rparagraph{Multi-turn vs.\ Single-turn.}
\citet{laban2025llms} suggest that LLMs can ``get lost'' in multi-turn conversations, performing worse than when all information is presented in a single turn. 
We investigate whether this phenomenon holds in our progressive information-seeking setting. 
For each turn $i$ in a dialogue $d$, we define two conditions:
\begin{enumerate}[topsep=2pt, itemsep=1pt]
    \item \textbf{Multi-turn:} The model is prompted with the full dialogue history up to that point, $h_{d,i}$, which includes the sequence of hints $\{q_{d,1}, a_{d,1}, \dots, q_{d,i-1}, a_{d,i-1}\}$ preceding the current query $q_{d,i}$.
    \item \textbf{Single-turn:} We create a single prompt containing a concise summary $S_{d,i}$ that synthesizes all information from the hints provided up to turn $i$.
\end{enumerate}
We then compare accuracy $z_{d,i}$ and confidence $c_{d,i}$ under both conditions. If models perform significantly better in the single-turn condition, it would suggest a cognitive burden in integrating information incrementally. 
Conversely, comparable or superior performance in the multi-turn setting would indicate that our structured, progressive framework effectively guides the model.

\begin{figure*}[t!]
    \centering    
    \includegraphics[width=0.93\linewidth]{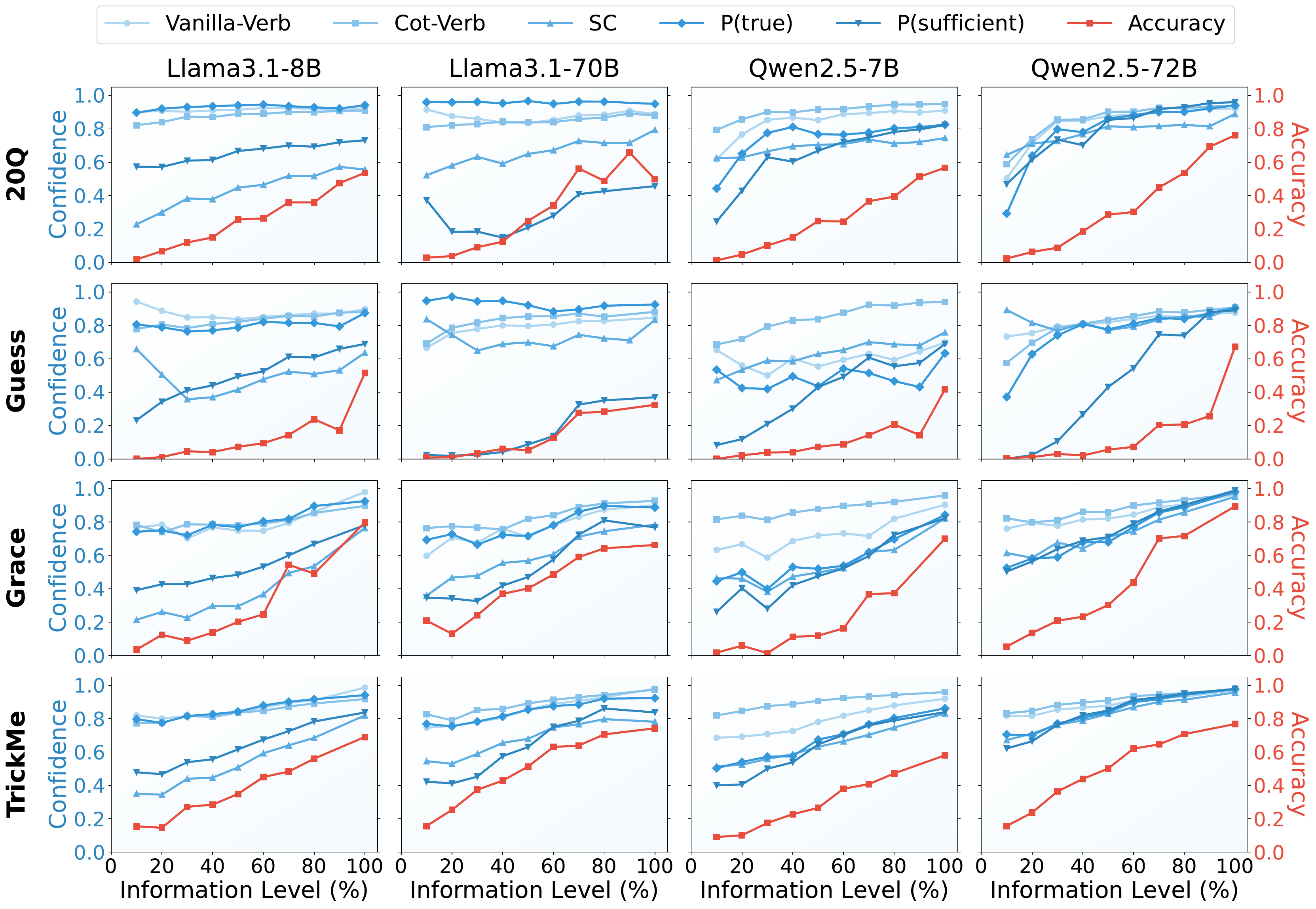}
    \caption{Evolution of average confidence and accuracy across different information levels. While accuracy (right y-axis, \textcolor{red}{red line}) generally increases, the confidence metrics (left y-axis, \textcolor{cyan}{blue line}) exhibit varying trends.}    
    \label{fig:information_gain}
\end{figure*}

\begin{figure*}[t!]
    \centering    \includegraphics[width=0.9\textwidth]{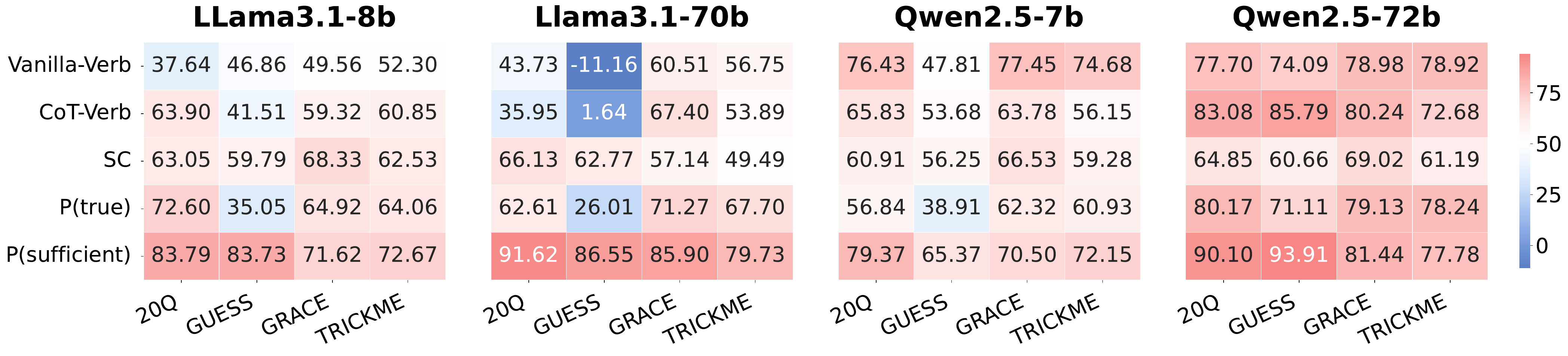}
    \caption{Kendall's $\tau$ for \textit{ground truth} answers. Compared to the $\tau$ for \textit{each turn's} answers, all methods show substantially better monotonicity. All values are shown as percentages.}    \label{fig:model_heatmaps}
\end{figure*}

\subsection{How reliable are confidence estimation methods in multi-turn settings?}
\label{sec:reliable-multiturn}

We assess reliability along two axes: 
\begin{inparaenum}[\it 1)]
    \item \emph{per-level calibration} using InfoECE (from 0 to 1, lower is better);
    \item \emph{monotonicity} of confidence over turns using Kendall’s $\tau$ (from -1 to 1, higher is better). We report $\tau$ both on the model’s \textbf{current} answer and on the \textbf{gold} answer at each turn (Table~\ref{tab:main}).
\end{inparaenum}
Figure~\ref{fig:information_gain} visualize how average confidence and accuracy evolve as information accumulates throughout the question-answering process. 


\rparagraph{Calibration is generally poor and sufficiency probes help the most.}
Across all models, both verbalized-based confidence (\textsc{Vanilla-Verb}, \textsc{CoT-Verb}) and logit-based \textsc{P(True)} are poorly calibrated, with InfoECE values typically between $40$ and $80$.
Self-consistency (\textsc{SC}) is usually the most calibrated on fully-specified incremental QA. 
In under-specified games, sufficiency probing can be strikingly better-calibrated: for Llama3.1-70B, InfoECE drops to $13.05$ on \twentyq\ and $5.27$ on \guess, while maintaining competitive performance on \grace\ and \trickme.
Overall, \textsc{SC} is a strong default for calibration. 
When the answer space is pruned gradually, \textsc{P(Sufficient)} is a more \textbf{\textit{efficient}} alternative. It narrows the gap and sometimes surpasses \textsc{SC}.

\rparagraph{Monotonicity on the current answer: sufficiency is usually best.}
Ideally, confidence should rise as clues accumulate. 
\textsc{P(Sufficient)} most consistently follows this trend: \eg $\tau=83.76$ on \guess with Qwen2.5-72B and $\tau=71.38$ on \trickme with Llama3.1-70B. 
In contrast, \textsc{SC} often shows weak monotonicity in under-specified settings (even single digits on \guess). 
There are model-family specific exceptions: with Qwen2.5 models, verbalized confidence (\textsc{Vanilla-Verb} or \textsc{CoT-Verb}) occasionally attains the highest $\tau$ scores on \twentyq\ and \grace, despite their generally poor calibration.

\rparagraph{Monotonicity on the ground truth: large gains and clear leaders.}
As shown in Figure \ref{fig:model_heatmaps}, when confidence is evaluated against the \textbf{ground truth} at each turn, all methods show substantial increases in $\tau$.
Although the ground truth is unavailable in real-world applications, this trend suggests that models can partially recognize when current hints align with the correct answer. 
\textsc{P(Sufficient)} dominates here, achieving $\tau = 93.91$ on \guess\ with Qwen2.5-72B, and $\tau = 91.62$, $86.55$, $85.90$ on \textsc{20Q}, \guess, and \grace\ with Llama3.1-70B, respectively. \textsc{Vanilla-Verb} can occasionally match or edge out on specific pairs (\eg Qwen2.5 on \textsc{20Q}), but it remains poorly calibrated.

\rparagraph{Scaling and Model Family Effects.} 
As parameters increase, we observe a consistent rise in accuracy and a marked improvement in $\tau$ (ranking calibration), particularly for the \textsc{P(Sufficient)}. For instance, \textbf{Qwen 2.5-72B} achieves the highest $\tau$ score of $83.76\%$ on the \textsc{Guess} dataset, significantly outperforming its 7B counterpart. However, the effect on \textsc{InfoECE} is more nuanced; while larger models generally provide more reliable rankings, smaller models occasionally exhibit lower absolute calibration errors in specific configurations. 


\begin{table*}[t]
\setlength\tabcolsep{13pt}
\centering
\scalebox{0.65}[0.65]{ 
\begin{tabular}{llAAACCC} 
\toprule
& \multirow{2}{*}{\textbf{Method}} 
& \multicolumn{3}{c}{\textbf{\twentyq}} 
& \multicolumn{3}{c}{\textbf{\guess}} \\
\cmidrule(lr){3-5} \cmidrule(lr){6-8}
& & \cellcolor{gray!0}$\rm{Conf}_{i-1}$ & \cellcolor{gray!0}$\rm{Conf}_{placebo, i}$ & \cellcolor{gray!0}$\rm{Conf}_{i}$ & \cellcolor{gray!0}$\rm{Conf}_{i-1}$ &\cellcolor{gray!0} $\rm{Conf}_{placebo, i}$ & \cellcolor{gray!0} $\rm{Conf}_{i}$ \\
\midrule
\multirow{5}{*}{\rotatebox{90}{\textbf{LLama3.1-8b}}}
& \textsc{Vanilla-Verb} & 90.69 & 88.42 \textcolor[HTML]{EB4869}{(2.27$\downarrow$)} & \underline{92.22} \textcolor[HTML]{5CC99C}{(1.53$\uparrow$)} & 87.87 & 89.54 \textcolor[HTML]{5CC99C}{(1.67$\uparrow$)} & 87.69 \textcolor[HTML]{EB4869}{(0.18$\downarrow$)} \\
& \textsc{Cot-Verb} & 85.20 & 85.18 \textcolor[HTML]{EB4869}{(0.02$\downarrow$)} & \underline{88.10} \textcolor[HTML]{5CC99C}{(2.90$\uparrow$)} & 79.26 & 81.32 \textcolor[HTML]{5CC99C}{(2.06$\uparrow$)} & \underline{84.26} \textcolor[HTML]{5CC99C}{(5.00$\uparrow$)} \\
& SC & 39.47 & \underline{44.29} \textcolor[HTML]{5CC99C}{(4.82$\uparrow$)} & 45.80 \textcolor[HTML]{5CC99C}{(6.33$\uparrow$)} & 37.22 & 34.13 \textcolor[HTML]{EB4869}{(3.09$\downarrow$)} & \underline{49.78} \textcolor[HTML]{5CC99C}{(12.56$\uparrow$)} \\
& \textsc{P(True)} & 91.51 & 89.74 \textcolor[HTML]{EB4869}{(1.77$\downarrow$)} & \underline{94.08} \textcolor[HTML]{5CC99C}{(2.57$\uparrow$)} & 73.63 & \underline{85.38} \textcolor[HTML]{5CC99C}{(11.75$\uparrow$)} & 83.32 \textcolor[HTML]{5CC99C}{(9.69$\uparrow$)} \\
& \textsc{P(Sufficient)} & 52.15 & \underline{49.21} \textcolor[HTML]{EB4869}{(2.94$\downarrow$)} & \underline{66.71} \textcolor[HTML]{5CC99C}{(14.56$\uparrow$)} & 44.18 & 45.88 \textcolor[HTML]{5CC99C}{(1.70$\uparrow$)} & \underline{54.02} \textcolor[HTML]{5CC99C}{(9.84$\uparrow$)} \\
\midrule
\multirow{5}{*}{\rotatebox{90}{\textbf{Llama3.1-70b}}}
& \textsc{Vanilla-Verb} & 87.30 & 84.03 \textcolor[HTML]{EB4869}{(3.27$\downarrow$)} & 85.84 \textcolor[HTML]{EB4869}{(1.46$\downarrow$)} & 71.30 & 73.70 \textcolor[HTML]{5CC99C}{(2.40$\uparrow$)} & \underline{83.83} \textcolor[HTML]{5CC99C}{(12.53$\uparrow$)} \\
& \textsc{Cot-Verb} & 85.11 & 84.29 \textcolor[HTML]{EB4869}{(0.82$\downarrow$)} & \underline{86.50} \textcolor[HTML]{5CC99C}{(1.39$\uparrow$)} & 78.39 & 78.77 \textcolor[HTML]{5CC99C}{(0.38$\uparrow$)} & \underline{88.41} \textcolor[HTML]{5CC99C}{(10.02$\uparrow$)} \\
& SC & 65.49 & 62.21 \textcolor[HTML]{EB4869}{(3.28$\downarrow$)} & 63.85 \textcolor[HTML]{EB4869}{(1.64$\downarrow$)} & 52.42 & 53.18 \textcolor[HTML]{5CC99C}{(0.76$\uparrow$)} & \underline{72.33} \textcolor[HTML]{5CC99C}{(19.91$\uparrow$)} \\
& \textsc{P(True)} & 95.48 & 93.43 \textcolor[HTML]{EB4869}{(2.05$\downarrow$)} & 94.56 \textcolor[HTML]{EB4869}{(0.92$\downarrow$)} & 88.16 & 88.14 \textcolor[HTML]{EB4869}{(0.02$\downarrow$)} & \underline{95.17} \textcolor[HTML]{5CC99C}{(7.01$\uparrow$)} \\
& \textsc{P(Sufficient)} & 19.95 & \underline{15.27} \textcolor[HTML]{EB4869}{(4.68$\downarrow$)} & \underline{33.29} \textcolor[HTML]{5CC99C}{(13.34$\uparrow$)} & 14.27 & \underline{2.97} \textcolor[HTML]{EB4869}{(11.30$\downarrow$)} & \underline{27.58} \textcolor[HTML]{5CC99C}{(13.31$\uparrow$)} \\
\midrule
\multirow{5}{*}{\rotatebox{90}{\textbf{Qwen2.5-7b}}}
& \textsc{Vanilla-Verb} & 86.31 & \underline{80.97} \textcolor[HTML]{EB4869}{(5.34$\downarrow$)} & \underline{86.97} \textcolor[HTML]{5CC99C}{(0.66$\uparrow$)} & 68.86 & 67.80 \textcolor[HTML]{EB4869}{(1.06$\downarrow$)} & \underline{59.57} \textcolor[HTML]{EB4869}{(9.29$\downarrow$)} \\
& \textsc{Cot-Verb} & 89.36 & \underline{85.13} \textcolor[HTML]{EB4869}{(4.23$\downarrow$)} & \underline{92.21} \textcolor[HTML]{5CC99C}{(2.85$\uparrow$)} & 81.48 & 82.96 \textcolor[HTML]{5CC99C}{(1.48$\uparrow$)} & 85.45 \textcolor[HTML]{5CC99C}{(3.97$\uparrow$)} \\
& SC & 66.73 & \underline{72.52} \textcolor[HTML]{5CC99C}{(5.79$\uparrow$)} & 69.69 \textcolor[HTML]{5CC99C}{(2.96$\uparrow$)} & 70.49 & \underline{65.74} \textcolor[HTML]{EB4869}{(4.75$\downarrow$)} & 65.87 \textcolor[HTML]{EB4869}{(4.62$\downarrow$)} \\
& \textsc{P(True)} & 81.17 & \underline{68.24} \textcolor[HTML]{EB4869}{(12.93$\downarrow$)} & \underline{77.24} \textcolor[HTML]{EB4869}{(3.93$\downarrow$)} & 39.69 & 35.91 \textcolor[HTML]{EB4869}{(3.78$\downarrow$)} & \underline{50.01} \textcolor[HTML]{5CC99C}{(10.32$\uparrow$)} \\
& \textsc{P(Sufficient)} & 46.66 & \underline{39.44} \textcolor[HTML]{EB4869}{(7.22$\downarrow$)} & \underline{67.58} \textcolor[HTML]{5CC99C}{(20.92$\uparrow$)} & 41.38 & \underline{20.04} \textcolor[HTML]{EB4869}{(21.34$\downarrow$)} & \underline{45.77} \textcolor[HTML]{5CC99C}{(4.39$\uparrow$)} \\
\midrule
\multirow{5}{*}{\rotatebox{90}{\textbf{Qwen2.5-72b}}}
& \textsc{Vanilla-Verb} & 77.90 & \underline{81.31} \textcolor[HTML]{5CC99C}{(3.41$\uparrow$)} & \underline{87.48} \textcolor[HTML]{5CC99C}{(9.58$\uparrow$)} & 76.35 & 73.07 \textcolor[HTML]{EB4869}{(3.28$\downarrow$)} & \underline{82.71} \textcolor[HTML]{5CC99C}{(6.36$\uparrow$)} \\
& \textsc{Cot-Verb} & 82.26 & \underline{84.42} \textcolor[HTML]{5CC99C}{(2.16$\uparrow$)} & \underline{88.63} \textcolor[HTML]{5CC99C}{(6.37$\uparrow$)} & 80.56 & \underline{75.74} \textcolor[HTML]{EB4869}{(4.82$\downarrow$)} & \underline{84.42} \textcolor[HTML]{5CC99C}{(3.86$\uparrow$)} \\
& SC & 80.71 & \underline{76.73} \textcolor[HTML]{EB4869}{(3.98$\downarrow$)} & 79.73 \textcolor[HTML]{EB4869}{(0.98$\downarrow$)} & 70.45 & 70.31 \textcolor[HTML]{EB4869}{(0.14$\downarrow$)} & \underline{81.93} \textcolor[HTML]{5CC99C}{(11.48$\uparrow$)} \\
& \textsc{P(True)} & 80.65 & 81.75 \textcolor[HTML]{5CC99C}{(1.10$\uparrow$)} & 85.01 \textcolor[HTML]{5CC99C}{(4.36$\uparrow$)} & 51.98 & \underline{66.59} \textcolor[HTML]{5CC99C}{(14.61$\uparrow$)} & \underline{81.25} \textcolor[HTML]{5CC99C}{(29.27$\uparrow$)} \\
& \textsc{P(Sufficient)} & 80.46 & 79.82 \textcolor[HTML]{EB4869}{(0.64$\downarrow$)} & 81.68 \textcolor[HTML]{5CC99C}{(1.22$\uparrow$)} & 64.96 & \underline{53.16} \textcolor[HTML]{EB4869}{(11.80$\downarrow$)} & 51.98 \textcolor[HTML]{EB4869}{(12.98$\downarrow$)} \\
\bottomrule
\end{tabular}
}
\caption{Average confidence comparison across different models and datasets. Values in parentheses show the change relative to $\rm{Conf}_{i-1}$. \underline{Underlined} values indicate statistically significant changes ($p<0.05$).}
\label{tab:vacauous}
\end{table*}

\subsection{Does confidence track information or just turn count?}
\label{sec:info-vs-length}
We test whether confidence increases are driven by accumulating information or are merely an artifact of conversational length. As detailed in our experimental setup and shown in Table~\ref{tab:vacauous}, we compare a baseline confidence (turn $i-1$) with the confidence after receiving either an \textit{informative} hint (Original) or a \textit{non-informative} placebo hint.
Across all 40 comparisons, informative turns yield more significant changes than placebos (27 vs.\ 18 with $p{<}0.05$). 
Among them, \textsc{P(Sufficient)} most cleanly disentangles information gain from mere turn accumulation.

\rparagraph{Sufficiency probes actively penalize uninformative turns, tracking evidence over length.}
The \textsc{P(Sufficient)} method proves to be the most robust by actively penalizing uninformative turns.
It frequently shows a statistically significant \textit{decrease} in confidence after a placebo hint (\eg a drop from 14.27 to 2.97 for Llama3.1-70B on \guess). 
This behavior, where the model lowers its sufficiency assessment in response to a useless hint, confirms it is tracking evidence, not just turn count. As a result, it achieves the clearest separation between conditions: confidence decreases or remains flat with a placebo, but increases with an informative hint. 
In contrast, other methods, particularly verbalized ones, can be misled into increasing confidence simply because the conversation has progressed.

\rparagraph{Placebo hints reveal important differences between methods.}
The adversarial condition with placebo hints shows that models are not simply becoming more confident as a conversation gets longer. For many methods, the change in confidence after a placebo hint is not statistically significant (high $p$-value). For example, for Llama3.1-70B on GUESS, \textsc{CoT-Verb}, \textsc{SC}, and \textsc{P(True)} show negligible changes ($p > 0.6$). This suggests a degree of robustness against superficial conversational structure.

\begin{figure*}[ht!]
    \centering    
    \includegraphics[width=0.92\textwidth]{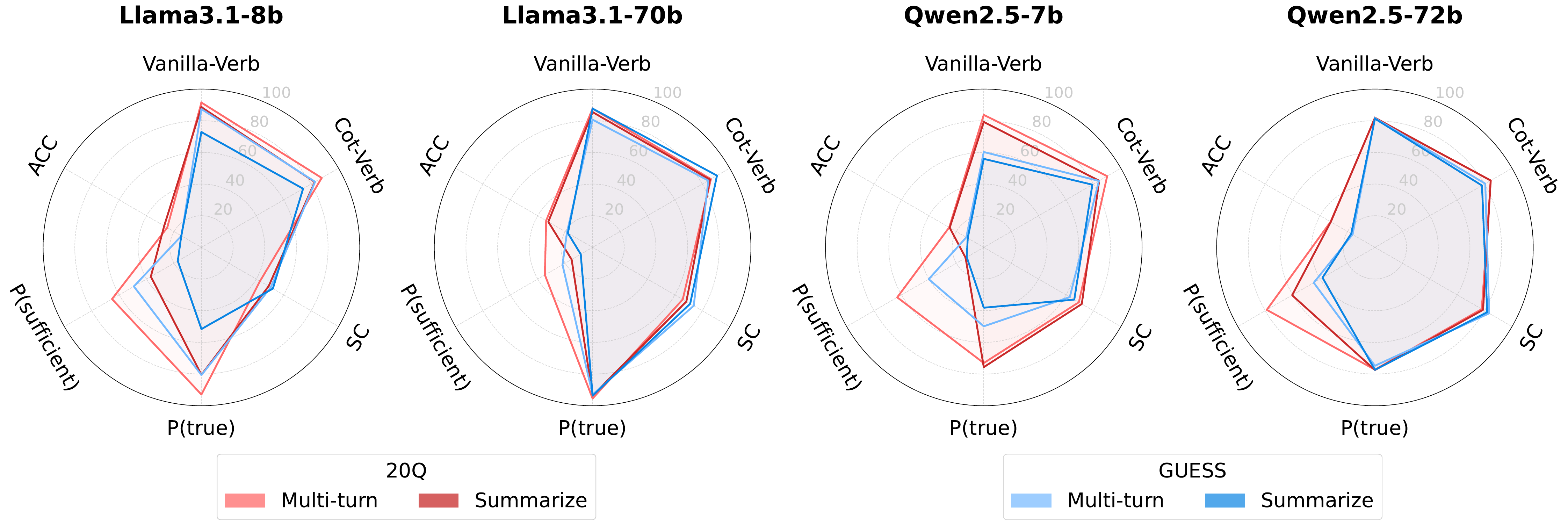}
    \caption{Performance comparison across four language models. Six evaluation dimensions are shown: Vanilla-Verb, Cot-Verb, SC, P(true), P(sufficient), and ACC. \textcolor{red}{Red} indicates 20Q benchmarks, \textcolor{blue}{blue} indicates GUESS benchmarks. }
    \vspace{-2mm}
    \label{fig:radar_charts}
\end{figure*}

\rparagraph{\textsc{P(True)} is confounded by turn count (especially in \guess).}
In open-ended \guess, \textsc{P(True)} often \emph{rises even under placebo}, consistent with a length artifact (mean $\Delta_{\text{placebo}}=+5.64$\,; significant in 2/4 model pairs). Notably, Llama3.1-8B jumps $+11.75$\, under placebo ($p{<}10^{-12}$) and Qwen2.5-72B jumps $+14.61$\, ($p{<}10^{-6}$). Although \textsc{P(True)} also increases with genuine information (mean $\Delta_{\text{info}}=+14.07$\, on \guess), the placebo lift makes it less reliable for disentangling information from dialogue length. On \twentyq, placebo effects tend to be negative (mean $-3.91$\,), suggesting format-dependent behavior.

\rparagraph{Self-consistency (\textsc{SC}) shows moderate robustness.}
\textsc{SC} typically exhibits small or negligible placebo movements and sizable gains with informative hints (mean $\Delta_{\text{info}}=+9.83$\, on \guess). However, it is \emph{not} immune: \eg Llama3.1-8B on \twentyq increases under placebo by $+4.82$\, ($p{=}0.0025$). Thus, \textsc{SC} generally tracks information better than verbalized scores but can still pick up turn-index artifacts in some settings.

\rparagraph{Verbalized confidence is unstable across conditions.}
\textsc{Vanilla-Verb}/\textsc{CoT-Verb} show small average placebo shifts (often non-significant) and modest informative gains; in some cases they even move counterintuitively (\eg Qwen2.5-7B on \guess decreases by $-9.29$\, with an informative hint, $p{=}0.005$). This instability, together with poor calibration (\S~\ref{sec:reliable-multiturn}), limits their utility for turn-by-turn reliability.

\subsection{Single-Turn Summary vs.\ Multi-Turn Interaction}
\label{sec:single-vs-multi}
We compare model behavior when consuming clues incrementally (\textbf{multi-turn}) versus reading a concise synthesis of the same clues in one prompt (\textbf{single-turn summary}). Across models and datasets, accuracy differences between the two settings are small (mean absolute gap ${<}1$\,), indicating no systematic advantage for either format (Figure~\ref{fig:radar_charts}). For example, on \twentyq Llama3.1-8B slightly improves with summaries ($24.95{\rightarrow}27.16$\,), whereas Llama3.1-70B slightly drops ($33.87{\rightarrow}32.31$\,). On \guess{}, Qwen2.5-72B gains modestly ($16.31{\rightarrow}17.29$\,), while Qwen2.5-7B loses ($12.74{\rightarrow}11.69$\,). In short, unlike the “getting lost” effect reported by \citet{laban2025llms}, our progressive information-seeking setup yields comparable task accuracy in multi-turn and single-turn conditions (see Appendix~\ref{sec:info-ece_summarize} for detailed InfoECE comparisons). One possible reason is that our tasks do not involve complicated arithmetic reasoning compared with \citet{laban2025llms}.

\rparagraph{Confidence shifts depend strongly on the signal.}
While accuracy is stable, confidence responds markedly to prompt format. The sufficiency probe \textsc{P(Sufficient)} consistently \emph{drops} under single-turn summaries (\eg on \twentyq: Qwen2.5-7B $63.13{\rightarrow}13.23$; Llama3.1-70B $34.80{\rightarrow}15.30$), suggesting that compressing the dialogue into a synopsis removes turn-structure cues that the probe exploits to assess whether evidence is \emph{enough}. \textsc{P(True)} and verbalized confidence often decrease with summaries for smaller models (\eg Llama3.1-8B on \guess: \textsc{P(True)} $80.66{\rightarrow}51.58$, \textsc{Vanilla-Verb} $87.33{\rightarrow}72.82$), but can \emph{increase} for larger models in some cases (\eg Llama3.1-70B on \guess: \textsc{Vanilla-Verb} $80.63{\rightarrow}87.65$, \textsc{CoT-Verb} $84.43{\rightarrow}90.72$) without commensurate accuracy gains—an instance of potential miscalibration inflation. By contrast, \textsc{SC} is comparatively stable and sometimes \emph{rises} with summaries on \twentyq{} (\eg Llama3.1-8B: $42.70{\rightarrow}49.04$), but shows mixed movement on \guess.


\section{Conclusion}

We present the first systematic study of confidence estimation for LLMs in multi-turn conversations. We establish a formal evaluation framework grounded in two key desiderata with novel metrics and datasets. 
Our evaluation across various confidence estimation methods reveals that widely-used techniques struggle to maintain calibration and monotonicity in dynamic dialogues. We find that our proposed logit-based probe, $P(\text{SUFFICIENT})$, achieves comparatively better performance; however, the task remains significantly under-resolved. 
Building on our foundation, we advocate for future research into methods that: \textbf{(1)} satisfy both calibration and monotonicity; \textbf{(2)} effectively distinguish task-relevant information from conversational filler; and \textbf{(3)} remain robust across both single-turn summaries and multi-turn interactions.

\section*{Limitation}

Our work, while providing a foundational framework, has several limitations that open avenues for future research. (1) The progressive datasets and Hinter–Guesser protocol simplify real conversations, omitting phenomena such as topic shifts, repairs, and mixed intents, which may limit transfer to messy, open-world dialogue. (2) Our study focuses on specific information-seeking tasks; the dynamics of confidence in more open-ended, creative, or collaborative conversations remain an open question. (3) Our evaluation emphasizes calibration and rank monotonicity; the downstream impact on user utility and human trust requires controlled user studies and field deployments. (4) We study \emph{confidence} rather than \emph{uncertainty}; extending our framework to uncertainty quantification and its relationship to confidence in multi-turn settings is an important next step.

\section*{Ethics Statement}

Our research follows standard ethical guidelines. We verified the licenses of all software and datasets used in this study.
We introduce new multi-turn evaluation datasets built from existing resources and automated generation. Despite best efforts, the data may contain Western-centric biases, and we did not address multilingual coverage, which may limit generality across languages and cultures.
Although some source datasets were created with human-in-the-loop protocols (\eg GRACE \citep{sung-etal-2025-grace}), our experiments are fully automated. We recruited no new participants or annotators; thus no compensation or IRB oversight was required.
We identified no privacy concerns, as we do not collect, store, or release personally identifiable information. We do not anticipate additional risks. We used an AI assistant only for grammar checking.

\section*{Acknowledgement}
We thank Michael Kirchhof for inspiring our work on confidence estimation in multi-turn settings and for the early discussions. We also thank the authors of \grace~\citep{sung-etal-2025-grace} and \trickme~\citep{wallace-etal-2019-trick} for providing the valuable datasets.
\bibliography{anthology,custom}

@misc{zhou2025beyond,
      title={Beyond the Final Layer: Intermediate Representations for Better Multilingual Calibration in Large Language Models}, 
      author={Ej Zhou and Caiqi Zhang and Tiancheng Hu and Chengzu Li and Nigel Collier and Ivan Vulić and Anna Korhonen},
      year={2025},
      eprint={2510.03136},
      archivePrefix={arXiv},
      primaryClass={cs.CL},
      url={https://arxiv.org/abs/2510.03136}, 
}

@misc{abdulhai2023lmrl,
    title={LMRL Gym: Benchmarks for Multi-Turn Reinforcement Learning with Language Models},
    author={Marwa Abdulhai and Isadora White and Charlie Snell and Charles Sun and Joey Hong and Yuexiang Zhai and Kelvin Xu and Sergey Levine},
    year={2023},
    eprint={2311.18232},
    archivePrefix={arXiv},
    primaryClass={cs.CL}
}

@article{llama3modelcard,
 author = {Meta},
 title = {Llama 3 Model Card},
 url = {https://github.com/meta-llama/llama3/blob/main/MODEL_CARD.md},
 year = {2024}
}

@misc{yang2024qwen2technicalreport,
 author = {An Yang and Baosong Yang and Binyuan Hui and Bo Zheng and Bowen Yu and Chang Zhou and Chengpeng Li and Chengyuan Li and Dayiheng Liu and Fei Huang and Guanting Dong and Haoran Wei and Huan Lin and Jialong Tang and Jialin Wang and Jian Yang and Jianhong Tu and Jianwei Zhang and Jianxin Ma and Jianxin Yang and Jin Xu and Jingren Zhou and Jinze Bai and Jinzheng He and Junyang Lin and Kai Dang and Keming Lu and Keqin Chen and Kexin Yang and Mei Li and Mingfeng Xue and Na Ni and Pei Zhang and Peng Wang and Ru Peng and Rui Men and Ruize Gao and Runji Lin and Shijie Wang and Shuai Bai and Sinan Tan and Tianhang Zhu and Tianhao Li and Tianyu Liu and Wenbin Ge and Xiaodong Deng and Xiaohuan Zhou and Xingzhang Ren and Xinyu Zhang and Xipin Wei and Xuancheng Ren and Xuejing Liu and Yang Fan and Yang Yao and Yichang Zhang and Yu Wan and Yunfei Chu and Yuqiong Liu and Zeyu Cui and Zhenru Zhang and Zhifang Guo and Zhihao Fan},
 journal = {ArXiv preprint},
 title = {Qwen2 Technical Report},
 url = {https://arxiv.org/abs/2407.10671},
 volume = {abs/2407.10671},
 year = {2024}
}

@misc{kadavath2022language,
 archiveprefix = {arXiv},
 author = {Saurav Kadavath and Tom Conerly and Amanda Askell and Tom Henighan and Dawn Drain and Ethan Perez and Nicholas Schiefer and Zac Hatfield-Dodds and Nova DasSarma and Eli Tran-Johnson and Scott Johnston and Sheer El-Showk and Andy Jones and Nelson Elhage and Tristan Hume and Anna Chen and Yuntao Bai and Sam Bowman and Stanislav Fort and Deep Ganguli and Danny Hernandez and Josh Jacobson and Jackson Kernion and Shauna Kravec and Liane Lovitt and Kamal Ndousse and Catherine Olsson and Sam Ringer and Dario Amodei and Tom Brown and Jack Clark and Nicholas Joseph and Ben Mann and Sam McCandlish and Chris Olah and Jared Kaplan},
 eprint = {2207.05221},
 primaryclass = {cs.CL},
 title = {Language Models (Mostly) Know What They Know},
 year = {2022}
}

@misc{zhang2024atomic,
 author = {Caiqi Zhang and Ruihan Yang and Zhisong Zhang and Xinting Huang and Sen Yang and Dong Yu and Nigel Collier},
 journal = {ArXiv preprint},
 title = {Atomic Calibration of LLMs in Long-Form Generations},
 url = {https://arxiv.org/abs/2410.13246},
 volume = {abs/2410.13246},
 year = {2024}
}

@inproceedings{xiong2023llms,
 author = {Miao Xiong and
Zhiyuan Hu and
Xinyang Lu and
Yifei Li and
Jie Fu and
Junxian He and
Bryan Hooi},
 bibsource = {dblp computer science bibliography, https://dblp.org},
 booktitle = {The Twelfth International Conference on Learning Representations,
{ICLR} 2024, Vienna, Austria, May 7-11, 2024},
 publisher = {OpenReview.net},
 title = {Can LLMs Express Their Uncertainty? An Empirical Evaluation of Confidence
Elicitation in LLMs},
 url = {https://openreview.net/forum?id=gjeQKFxFpZ},
 year = {2024}
}

@misc{lin2023generating,
 archiveprefix = {arXiv},
 author = {Zhen Lin and Shubhendu Trivedi and Jimeng Sun},
 eprint = {2305.19187},
 primaryclass = {cs.CL},
 title = {Generating with Confidence: Uncertainty Quantification for Black-box Large Language Models},
 year = {2023}
}

@misc{zhang2025reinforcement,
    title={Reinforcement Learning for Better Verbalized Confidence in Long-Form Generation},
    author={Caiqi Zhang and Xiaochen Zhu and Chengzu Li and Nigel Collier and Andreas Vlachos},
    year={2025},
    eprint={2505.23912},
    archivePrefix={arXiv},
    primaryClass={cs.CL}
}

@inproceedings{yang-etal-2025-logu,
    title = "{L}o{GU}: Long-form Generation with Uncertainty Expressions",
    author = "Yang, Ruihan  and
      Zhang, Caiqi  and
      Zhang, Zhisong  and
      Huang, Xinting  and
      Yang, Sen  and
      Collier, Nigel  and
      Yu, Dong  and
      Yang, Deqing",
    editor = "Che, Wanxiang  and
      Nabende, Joyce  and
      Shutova, Ekaterina  and
      Pilehvar, Mohammad Taher",
    booktitle = "Proceedings of the 63rd Annual Meeting of the Association for Computational Linguistics (Volume 1: Long Papers)",
    month = jul,
    year = "2025",
    address = "Vienna, Austria",
    publisher = "Association for Computational Linguistics",
    url = "https://aclanthology.org/2025.acl-long.928/",
    doi = "10.18653/v1/2025.acl-long.928",
    pages = "18947--18968",
    ISBN = "979-8-89176-251-0",
}

@inproceedings{shelmanov-etal-2025-head,
    title = "A Head to Predict and a Head to Question: Pre-trained Uncertainty Quantification Heads for Hallucination Detection in {LLM} Outputs",
    author = "Shelmanov, Artem  and
      Fadeeva, Ekaterina  and
      Tsvigun, Akim  and
      Tsvigun, Ivan  and
      Xie, Zhuohan  and
      Kiselev, Igor  and
      Daheim, Nico  and
      Zhang, Caiqi  and
      Vazhentsev, Artem  and
      Sachan, Mrinmaya  and
      Nakov, Preslav  and
      Baldwin, Timothy",
    editor = "Christodoulopoulos, Christos  and
      Chakraborty, Tanmoy  and
      Rose, Carolyn  and
      Peng, Violet",
    booktitle = "Proceedings of the 2025 Conference on Empirical Methods in Natural Language Processing",
    month = nov,
    year = "2025",
    address = "Suzhou, China",
    publisher = "Association for Computational Linguistics",
    url = "https://aclanthology.org/2025.emnlp-main.1809/",
    doi = "10.18653/v1/2025.emnlp-main.1809",
    pages = "35700--35719",
    ISBN = "979-8-89176-332-6",
}

@article{zhang2026confidence,
  title={Confidence Should Be Calibrated More Than One Turn Deep},
  author={Zhang, Zhaohan and Li, Chengzhengxu and Liu, Xiaoming and Shen, Chao and Liu, Ziquan and Patras, Ioannis},
  journal={arXiv preprint arXiv:2604.05397},
  year={2026}
}

@article{zhang2025grace,
  title={Grace: A generative approach to better confidence elicitation in large language models},
  author={Zhang, Zhaohan and Liu, Ziquan and Patras, Ioannis},
  journal={arXiv preprint arXiv:2509.09438},
  year={2025}
}

@inproceedings{yang-etal-2025-uncle,
    title = "{UNCLE}: Benchmarking Uncertainty Expressions in Long-Form Generation",
    author = "Yang, Ruihan  and
      Zhang, Caiqi  and
      Zhang, Zhisong  and
      Huang, Xinting  and
      Yu, Dong  and
      Collier, Nigel  and
      Yang, Deqing",
    editor = "Christodoulopoulos, Christos  and
      Chakraborty, Tanmoy  and
      Rose, Carolyn  and
      Peng, Violet",
    booktitle = "Proceedings of the 2025 Conference on Empirical Methods in Natural Language Processing",
    month = nov,
    year = "2025",
    address = "Suzhou, China",
    publisher = "Association for Computational Linguistics",
    url = "https://aclanthology.org/2025.emnlp-main.1543/",
    doi = "10.18653/v1/2025.emnlp-main.1543",
    pages = "30328--30344",
    ISBN = "979-8-89176-332-6",
}

@inproceedings{zhu-etal-2025-conformity,
    title = "Conformity in Large Language Models",
    author = "Zhu, Xiaochen  and
      Zhang, Caiqi  and
      Stafford, Tom  and
      Collier, Nigel  and
      Vlachos, Andreas",
    editor = "Che, Wanxiang  and
      Nabende, Joyce  and
      Shutova, Ekaterina  and
      Pilehvar, Mohammad Taher",
    booktitle = "Proceedings of the 63rd Annual Meeting of the Association for Computational Linguistics (Volume 1: Long Papers)",
    month = jul,
    year = "2025",
    address = "Vienna, Austria",
    publisher = "Association for Computational Linguistics",
    url = "https://aclanthology.org/2025.acl-long.195/",
    doi = "10.18653/v1/2025.acl-long.195",
    pages = "3854--3872",
    ISBN = "979-8-89176-251-0",
}

@inproceedings{zhang2025roads,
    title = "All Roads Lead to {R}ome: Graph-Based Confidence Estimation for Large Language Model Reasoning",
    author = "Zhang, Caiqi  and
      Shu, Chang  and
      Shareghi, Ehsan  and
      Collier, Nigel",
    editor = "Christodoulopoulos, Christos  and
      Chakraborty, Tanmoy  and
      Rose, Carolyn  and
      Peng, Violet",
    booktitle = "Proceedings of the 2025 Conference on Empirical Methods in Natural Language Processing",
    month = nov,
    year = "2025",
    address = "Suzhou, China",
    publisher = "Association for Computational Linguistics",
    url = "https://aclanthology.org/2025.emnlp-main.1620/",
    doi = "10.18653/v1/2025.emnlp-main.1620",
    pages = "31802--31812",
    ISBN = "979-8-89176-332-6",
}

@inproceedings{zhang-zhang-2025-cot,
    title = "{C}o{T}-{UQ}: Improving Response-wise Uncertainty Quantification in {LLM}s with Chain-of-Thought",
    author = "Zhang, Boxuan  and
      Zhang, Ruqi",
    editor = "Che, Wanxiang  and
      Nabende, Joyce  and
      Shutova, Ekaterina  and
      Pilehvar, Mohammad Taher",
    booktitle = "Findings of the Association for Computational Linguistics: ACL 2025",
    month = jul,
    year = "2025",
    address = "Vienna, Austria",
    publisher = "Association for Computational Linguistics",
    url = "https://aclanthology.org/2025.findings-acl.1339/",
    doi = "10.18653/v1/2025.findings-acl.1339",
    pages = "26114--26133",
    ISBN = "979-8-89176-256-5",
}

@misc{laban2025llms,
    title={LLMs Get Lost In Multi-Turn Conversation},
    author={Philippe Laban and Hiroaki Hayashi and Yingbo Zhou and Jennifer Neville},
    year={2025},
    eprint={2505.06120},
    archivePrefix={arXiv},
    primaryClass={cs.CL}
}

@misc{yi2024survey,
    title={A Survey on Recent Advances in LLM-Based Multi-turn Dialogue Systems},
    author={Zihao Yi and Jiarui Ouyang and Zhe Xu and Yuwen Liu and Tianhao Liao and Haohao Luo and Ying Shen},
    year={2024},
    eprint={2402.18013},
    archivePrefix={arXiv},
    primaryClass={cs.CL}
}

@misc{wang2023mint,
    title={MINT: Evaluating LLMs in Multi-turn Interaction with Tools and Language Feedback},
    author={Xingyao Wang and Zihan Wang and Jiateng Liu and Yangyi Chen and Lifan Yuan and Hao Peng and Heng Ji},
    year={2023},
    eprint={2309.10691},
    archivePrefix={arXiv},
    primaryClass={cs.CL}
}

@inproceedings{sung-etal-2025-grace,
    title = "{GRACE}: A Granular Benchmark for Evaluating Model Calibration against Human Calibration",
    author = "Sung, Yoo Yeon  and
      Fleisig, Eve  and
      Hou, Yu  and
      Upadhyay, Ishan  and
      Boyd-Graber, Jordan Lee",
    editor = "Che, Wanxiang  and
      Nabende, Joyce  and
      Shutova, Ekaterina  and
      Pilehvar, Mohammad Taher",
    booktitle = "Proceedings of the 63rd Annual Meeting of the Association for Computational Linguistics (Volume 1: Long Papers)",
    month = jul,
    year = "2025",
    address = "Vienna, Austria",
    publisher = "Association for Computational Linguistics",
    url = "https://aclanthology.org/2025.acl-long.962/",
    doi = "10.18653/v1/2025.acl-long.962",
    pages = "19586--19587",
    ISBN = "979-8-89176-251-0",
}

@misc{kirchhof2025position,
    title={Position: Uncertainty Quantification Needs Reassessment for Large-language Model Agents},
    author={Michael Kirchhof and Gjergji Kasneci and Enkelejda Kasneci},
    year={2025},
    eprint={2505.22655},
    archivePrefix={arXiv},
    primaryClass={cs.LG}
}

@article{manakul2023selfcheckgpt,
  title={Selfcheckgpt: Zero-resource black-box hallucination detection for generative large language models},
  author={Manakul, Potsawee and Liusie, Adian and Gales, Mark JF},
  journal={arXiv preprint arXiv:2303.08896},
  year={2023}
}

@article{hu2025navigating,
  title={Navigating the Alignment-Calibration Trade-off: A Pareto-Superior Frontier via Model Merging},
  author={Hu, Tiancheng and Minixhofer, Benjamin and Collier, Nigel},
  journal={arXiv preprint arXiv:2510.17426},
  year={2025}
}
\clearpage
\appendix
\section*{Appendix}
\label{sec:appendix}

\section{Instruction Prompt Examples.}
\label{app:ambig_prompt}
\begin{table}[h]
\small
  \centering
    \begin{tabularx}{\linewidth}{X}
    \toprule
    \rowcolor[gray]{0.95}\multicolumn{1}{c}{\textbf{I: GENERATION TEMPLATE}} \\
    \midrule
    \{Question\} \\
    
    Use \textbackslash boxed\{\} to wrap your answer; for example \textbackslash boxed\{\{a concise answer\}\}. Your answer is: \\
    
    \midrule
    \rowcolor[gray]{0.95}\multicolumn{1}{c}{\textbf{II: VANILLA-VERB TEMPLATE}} \\
    \midrule
    \{Question\} \\
    
    Proposed Answer: \{Answer\} \\
    How confident are you that the proposed answer is correct? \\
    
    The confidence score should be a number from 0 (completely unsure) to 100 (absolutely certain). \\
    Your response MUST strictly adhere to this format: \\
    \#\#\# Confidence: \textbackslash boxed\{Your confidence score from 0-100.\} \\
    Your response is: \\
    
    \midrule
    \rowcolor[gray]{0.95}\multicolumn{1}{c}{\textbf{III: COT-VERB TEMPLATE}} \\
    \midrule
    \{Question\} \\
    Proposed Answer: \{Answer\} \\
    How confident are you that the proposed answer is correct? \\
    
    Analyze first and think step by step. The confidence score should be a number from 0 (completely unsure) to 100 (absolutely certain). \\
    
    Your response MUST strictly adhere to this format: \\
    \#\#\# Analysis: \textbackslash boxed\{Analyze the question and guess the answer. Think step by step and justify your confidence score.\} \\
    \#\#\# Confidence: \textbackslash boxed\{Your confidence score from 0-100.\} \\
    
    Your response is: \\
    \bottomrule
    \end{tabularx}
  \caption{Three TEMPLATE formats: generation and verbalize confidence estimation.}
  \label{tab:query_verb}
\end{table}
\begin{table}[h]
\small
  \centering
    \begin{tabularx}{\linewidth}{X}
    \toprule
    \rowcolor[gray]{0.95}\multicolumn{1}{c}{\textbf{I: P(Sufficient) TEMPLATE}} \\
    \midrule
   \{Question\}

    Based only on the information and hints provided above, does that information sufficiently entail that the correct answer is exactly \{Answer\}? \\
    
    A. Yes — the information is sufficient to conclude \{Answer\}. \\
    B. No — the information is insufficient, allows alternatives, or contradicts \{Answer\}. \\
    
    Output format: **A** or **B** only (single uppercase letter; no spaces, punctuation, or explanation): \\
    
    \midrule

    \rowcolor[gray]{0.95}\multicolumn{1}{c}{\textbf{II: P(Ture) TEMPLATE}} \\

    \midrule

    \{Question\} 
    
    Proposed Answer: \{Answer\}

    Is the proposed answer: \\
        A. True \\
        B. False \\
    Output format: **A** or **B** only (single uppercase letter; no spaces, punctuation, or explanation): \\
    \bottomrule    
    \end{tabularx}
  \caption{Two PROMPT\_TEMPLATE formats for binary classification tasks.}
  \label{tab:prompt_templates}
\end{table}
\begin{table}[h]
\small
  \centering
    \begin{tabularx}{\linewidth}{X}
    \toprule
    \rowcolor[gray]{0.95}\multicolumn{1}{c}{\textbf{SUMMARIZE PROMPT TEMPLATE}} \\
    \midrule
    You are given a block of hints about an entity (e.g., from a 20 Questions game). \\
    
    Summarize the hints into a concise statement or two, keeping the essential information instead of the Question/Answer format. \\
    
    Do not add new information or remove key attributes. \\
    
    Keep it as short and factual as possible. \\
    \bottomrule    
    \end{tabularx}
  \caption{SUMMARIZE\_PROMPT template for converting Q\&A format hints into concise factual statements.}
  \label{tab:summarize_prompt}
\end{table}

\section{Incremental QA datasets}
\label{app:incremental_qa}
\textbf{GRACE}~\citep{sung-etal-2025-grace}: a granular benchmark composed of quizbowl-style questions in which clues become increasingly specific. Each clue is self-contained and unambiguous. Models are evaluated on \emph{how early}, \emph{how accurately}, and \emph{how confidently} they answer as clues unfold, providing fine-grained signals for step-wise calibration. This directly supports \textbf{C1} and \textbf{C2}, and the pyramidal clue structure encourages \textbf{C3} as evidence strengthens.

\textbf{TrickMe}~\citep{wallace-etal-2019-trick}: a human-in-the-loop, adversarially-authored QA dataset built also in a quizbowl interface. Writers iteratively craft incremental clues to elicit confident model mistakes while remaining solvable by humans, yielding challenging, diverse questions that reveal miscalibration under partial information. It uses exactly the same task format as GRACE thus also satisfy our three criteria.

Together, GRACE and TrickMe instantiate the fully-specified setting where a unique gold answer exists from the outset, but models must calibrate \emph{when} to commit as evidence accrues. We use both datasets without altering their underlying incremental-clue protocols, and we report per-turn accuracy and confidence to align with our evaluation framework.

\section{Calibration shifts reveal a scaling-dependent format effect.}
\label{sec:info-ece_summarize}
\begin{figure*}[ht!]
    \centering        \includegraphics[width=\textwidth]{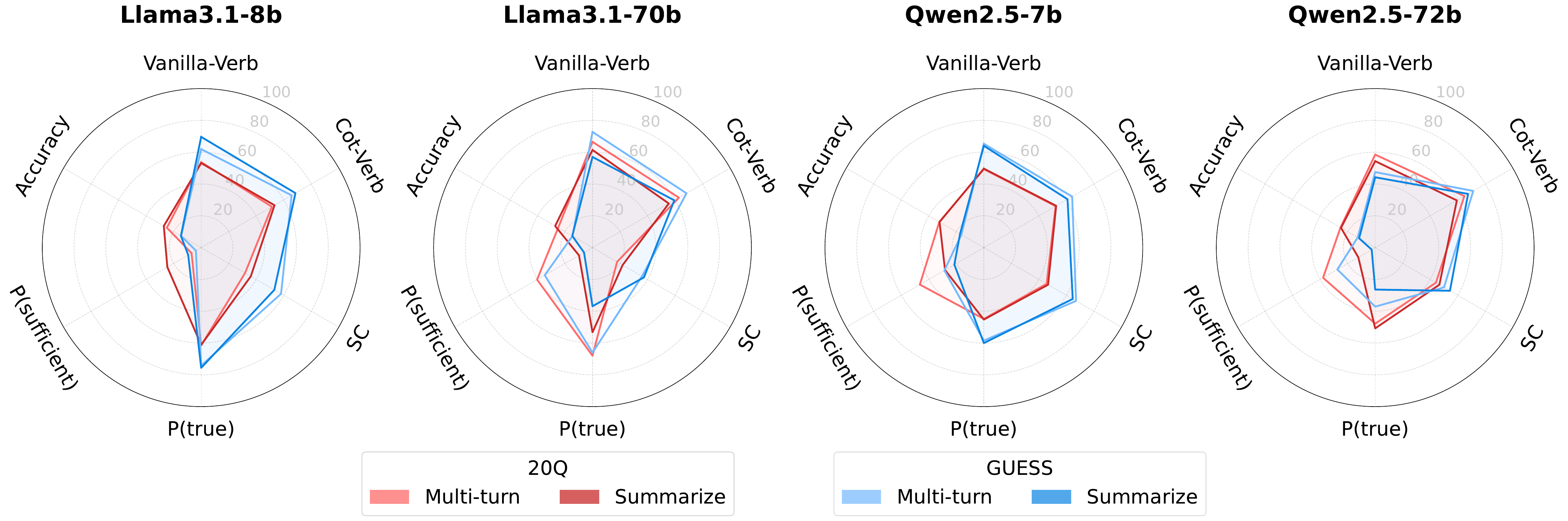}
    \caption{InfoECE of different confidence estimation methods across formats. Five confidence methods (Vanilla-Verb, CoT-Verb, SC, P(true), P(sufficient)) plus Accuracy, compared under multi-turn (darker) vs.\ summarized (lighter) presentation. \textcolor{red}{Red}: \twentyq; \textcolor{blue}{blue}: \guess. Lower InfoECE indicates better calibration.}
    \vspace{-2mm}
    \label{fig:ece_radar_charts}
\end{figure*}

While accuracy remains comparable across formats (\S\ref{sec:single-vs-multi}), calibration quality—measured by InfoECE—responds divergently by model scale (Figure~\ref{fig:ece_radar_charts}). The sufficiency probe \textsc{P(Sufficient)} exhibits opposite trends: for smaller models, summarization \emph{degrades} calibration (\eg Llama3.1-8B on \twentyq: $6.99{\rightarrow}24.57$; on \guess: $3.82{\rightarrow}9.41$), suggesting reliance on turn-by-turn structure. In stark contrast, larger models show substantial \emph{improvements} under summarization (\eg Llama3.1-70B on \twentyq: $40.29{\rightarrow}9.81$; on \guess: $34.70{\rightarrow}6.16$; Qwen2.5-72B on \guess: $27.51{\rightarrow}2.55$), indicating more effective integration of compressed evidence. \textsc{P(True)} improves markedly for Llama3.1-70B (\twentyq: $68.00{\rightarrow}53.28$; \guess: $66.07{\rightarrow}36.87$) but shows minimal change for smaller models. \textsc{SC} and verbalized methods remain largely format-invariant (shifts typically ${<}5$ InfoECE points) but consistently poorly calibrated (${>}50$ InfoECE). This scaling-dependent divergence suggests that while smaller models depend on conversational structure for reliable calibration, larger models can flexibly exploit either format, sometimes achieving superior calibration from summarization.

\section{Question Examples}
\begin{table*}[t]
\setlength\tabcolsep{11pt}
\footnotesize
  \centering
    \begin{tabularx}{\linewidth}{p{1cm}|X}
    \toprule
    \textbf{Dataset} & \textbf{Prompts} \\
    \midrule
    \multirow{7}{*}{\textbf{\twentyq}} & \textbf{User:} Given the following information, provide the title of the Wikipedia page that best answers the last question fragment. If unsure, provide your best guess. The answer should be concise. \\
    & You have some clues about the answer: \\
    & \textbf{Assistant:} Is it human-made? \textbf{User:} Yes \\
    & \textbf{Assistant:} Is it typically found indoors? \textbf{User:} Yes \\
    & \textbf{Assistant:} Is it commonly encountered in living rooms? \textbf{User:} Yes \\
    & \textbf{Assistant:} Is it larger than a book? \textbf{User:} Yes \\
    & \textbf{User:} Now guess what it is: \\
    & \textit{Keyword:} \colorbox{blue!15}{\textcolor{blue}{\textbf{television}}} \\
    \midrule
    \multirow{7}{*}{\parbox{2cm}{\textbf{\guess}}} & \textbf{User:} Given the following information, name the single CITY that best fits them. If unsure, provide your best guess. The answer should be concise. \\
    & You have some clues about the answer: \\
    & \textbf{Assistant:} What continent is the city in? \textbf{User:} Asia \\
    & \textbf{Assistant:} Is the city coastal or inland? \textbf{User:} Inland \\
    & \textbf{Assistant:} What's the climate like in the city? \textbf{User:} Tropical \\
    & \textbf{Assistant:} What region within the continent is the city located? \textbf{User:} Southeast Asia \\
    & \textbf{User:} Now guess what it is: \\
    & \textit{Keyword:} \colorbox{blue!15}{\textcolor{blue}{\textbf{Bogor, Indonesia}}} \\
    \midrule
    \multirow{8}{*}{\textbf{\grace}} & \textbf{User:} Given the following information, provide the title of the Wikipedia page that best answers the last question fragment. If unsure, provide your best guess. The answer should be concise. \\
    & You have some clues about the answer: \\
    & \textbf{User:} It's not Charlie Parker, but a musician with this surname arranged excerpts from Stravinsky's The Firebird and the "Goin' Home" theme from the New World Symphony. \\
    & \textbf{User:} Another album by a musician with this surname features chants from the Bhagavad Gita and solos by Pharoah Sanders. \\
    & \textbf{User:} A song by a musician with this surname features Elvin Jones on timpani and gong. \\
    & \textbf{User:} A set of chord substitutions in a ii-V-I ("two-five-one") progression that proved challenging for pianist Tommy Flanagan are called this surname's "changes." \\
    & \textbf{User:} For 10 points, give this surname of harpist Alice and her husband, the saxophonist behind the album Giant Steps. \\
    & \textbf{User:} Now guess what it is: \\
    & \textit{Keyword:} \colorbox{blue!15}{\textcolor{blue}{\textbf{Coltrane}}} \\
    \midrule
    \multirow{7}{*}{\textbf{\trickme}} & \textbf{User:} Given the following information, provide the title of the Wikipedia page that best answers the last question fragment. If unsure, provide your best guess. The answer should be concise. \\
    & You have some clues about the answer: \\
    & \textbf{User:} This man was seen driving Desiigner in the music video for Panda. \\
    & \textbf{User:} In an interview with Sway, this man yelled "I am Warhol," and compared himself with Shakespeare. \\
    & \textbf{User:} This man also controversially said that slavery was a choice. \\
    & \textbf{User:} For 10 points, name this songwriter known for his songs "Power," and "Gold Diggers," and more recently, "I Love It" with Lil Pump. \\
    & \textbf{User:} Now guess what it is: \\
    & \textit{Keyword:} \colorbox{blue!15}{\textcolor{blue}{\textbf{Kanye\_West}}} \\
    \bottomrule
    \end{tabularx}
  \caption{Examples from four datasets, showing question-answer dialogue format.}
  \label{tab:complete_prompts}
\end{table*}
We list some examples of the four datasets we use in the study in Table \ref{tab:complete_prompts}.

\section{Placebo QA Examples}
\begin{table*}[htbp]
\scriptsize
  \centering
    \begin{tabularx}{0.95\linewidth}{X|X}
    \toprule
    \textbf{Guess My City} & \textbf{20Q} \\
    \midrule
    \textbf{Q:} Does the city have people living in it? \textbf{A:} Yes & 
    \textbf{Q:} Can it be described using words? \textbf{A:} Yes \\
    
    \textbf{Q:} Does the city contain buildings? \textbf{A:} Yes & 
    \textbf{Q:} Can people ask questions about it? \textbf{A:} Yes \\
    
    \textbf{Q:} Are there roads in the city? \textbf{A:} Yes & 
    \textbf{Q:} Could it, in principle, be identified or referred to? \textbf{A:} Yes \\
    
    \textbf{Q:} Does the city have some form of waste disposal, like bins or trash collection? \textbf{A:} Yes & 
    \textbf{Q:} Does it have at least one property? \textbf{A:} Yes \\
    
    \textbf{Q:} Is there access to toilets in the city? \textbf{A:} Yes & 
    \textbf{Q:} Is it what it is? \textbf{A:} Yes \\
    
    \textbf{Q:} Does the city have shops or markets? \textbf{A:} Yes & 
    \textbf{Q:} Could someone think about it? \textbf{A:} Yes \\
    
    \textbf{Q:} Are there schools or educational institutions in the city? \textbf{A:} Yes & 
    \textbf{Q:} Can it be distinguished from nothing at all? \textbf{A:} Yes \\
    
    \textbf{Q:} Does the city have hospitals or clinics? \textbf{A:} Yes & 
    \textbf{Q:} Is it possible to talk about it right now? \textbf{A:} Yes \\
    
    \textbf{Q:} Is there some form of public transportation in the city? \textbf{A:} Yes & 
    \textbf{Q:} Does it have some kind of name or label? \textbf{A:} Yes \\
    
    \textbf{Q:} Does the city have restaurants or places to eat? \textbf{A:} Yes & 
    \textbf{Q:} Would it still count as something even if we know little about it? \textbf{A:} Yes \\
    
    \textbf{Q:} Are there offices or workplaces in the city? \textbf{A:} Yes & 
    \textbf{Q:} Could it, in theory, be observed or detected? \textbf{A:} Yes \\
    
    \textbf{Q:} Does the city have places for recreation, such as parks or sports areas? \textbf{A:} Yes & 
    \textbf{Q:} Does it interact with its environment in some way? \textbf{A:} Yes \\
    
    \textbf{Q:} Is the city located on land? \textbf{A:} Yes & 
    \textbf{Q:} Could it be distinguished from absolutely nothing? \textbf{A:} Yes \\
    
    \textbf{Q:} Does the city belong to a country? \textbf{A:} Yes & 
    \textbf{Q:} Is it possible to classify it as something rather than nothing? \textbf{A:} Yes \\
    
    \textbf{Q:} Is there some form of government or administration in the city? \textbf{A:} Yes & 
    \textbf{Q:} Does it occupy some kind of location, even if unknown? \textbf{A:} Yes \\
    
    \textbf{Q:} Does the city have streets or pathways for movement? \textbf{A:} Yes & 
    \textbf{Q:} Is it part of reality? \textbf{A:} Yes \\
    
    \textbf{Q:} Are there people who work in the city? \textbf{A:} Yes & 
    \textbf{Q:} Does it have some relation to other things? \textbf{A:} Yes \\
    
    \textbf{Q:} Does the city have some form of shelter or housing? \textbf{A:} Yes & 
    \textbf{Q:} Could one imagine it being measured somehow? \textbf{A:} Yes \\
    
    \textbf{Q:} Is there electricity available in parts of the city? \textbf{A:} Yes & 
    \textbf{Q:} Is Earth around the Sun? \textbf{A:} Yes \\
    
    \textbf{Q:} Does the city have some form of water supply or access? \textbf{A:} Yes & 
    \textbf{Q:} Is the Moon larger than the Sun? \textbf{A:} No \\
    
    \textbf{Q:} Are there vehicles that operate in the city? \textbf{A:} Yes & 
    \textbf{Q:} Can numbers be even? \textbf{A:} Yes \\
    
    \textbf{Q:} Does the city have some form of communication infrastructure? \textbf{A:} Yes & 
    \textbf{Q:} Is blue a kind of sound? \textbf{A:} No \\
    
    \textbf{Q:} Are there businesses operating in the city? \textbf{A:} Yes & 
    \textbf{Q:} Can a thought have weight? \textbf{A:} No \\
    
    \textbf{Q:} Does the city have some form of lighting at night? \textbf{A:} Yes & 
    \textbf{Q:} Is time measured by clocks? \textbf{A:} Yes \\
    
    \textbf{Q:} Are there emergency services available in the city? \textbf{A:} Yes & 
    \textbf{Q:} Do triangles have three sides? \textbf{A:} Yes \\
    
    \textbf{Q:} Does the city have banking or financial services? \textbf{A:} Yes & 
    \textbf{Q:} Is water wetter than fire? \textbf{A:} Yes \\
    
    \textbf{Q:} Are there entertainment venues in the city? \textbf{A:} Yes & \\
    
    \textbf{Q:} Does the city have postal or delivery services? \textbf{A:} Yes & \\
    
    \textbf{Q:} Are there religious or cultural institutions in the city? \textbf{A:} Yes & \\
    
    \textbf{Q:} Does the city have some form of law enforcement? \textbf{A:} Yes & \\
    
    \bottomrule
    \end{tabularx}
  \caption{We construct placebo question–answer pairs for control experiments using Guess My City (30 questions) and 20Q (26 questions). These questions neither provide any useful information to answer the question nor contradict with conversation history.}
  \label{tab:vacuous_qa}
\end{table*}
We list the placebo QA examples we use for \twentyq and \guess datasets in Table \ref{tab:vacuous_qa}.
\end{document}